\documentclass{article}

\usepackage{arxiv}

\usepackage[utf8]{inputenc} 
\usepackage[T1]{fontenc}    
\usepackage{hyperref}       
\usepackage{url}            
\usepackage{booktabs}       
\usepackage{amsfonts}       
\usepackage{nicefrac}       
\usepackage{microtype}      
\usepackage{lipsum}
\usepackage{graphicx}
\graphicspath{ {./images/} }

\usepackage{multirow}
\usepackage{subfigure}
\usepackage{enumitem}

\usepackage{color}
\definecolor{darkgreen}{rgb}{0,0.55,0}

\begin{document}

\title{Assessing the quality of sources in Wikidata across languages: a hybrid approach}

\author{
 Gabriel Amaral \\
  King's College London\\
  London WC2R 2LS, UK \\
  \texttt{gabriel.amaral@kcl.ac.uk} \\
   \And
 Alessandro Piscopo \\
  BBC\\
  London W12 7TQ \\
  \texttt{alessandro.piscopo@bbc.co.uk} \\
  \And
 Lucie-Aimée Kaffee \\
  University of Southampton\\
  Southampton SO17 1BJ, UK \\
  \texttt{kaffee@soton.ac.uk} \\
    \And
 Odinaldo Rodrigues \\
  King's College London\\
  London WC2R 2LS, UK \\
  \texttt{odinaldo.rodrigues@kcl.ac.uk} \\
    \And
 Elena Simperl \\
  King's College London\\
  London WC2R 2LS, UK \\
  \texttt{elena.simperl@kcl.ac.uk} \\
}


\begin{abstract}

Wikidata is one of the most important sources of structured data on the web, built by a worldwide community of volunteers.
As a secondary source, its contents must be backed by credible references; 
this is particularly important as Wikidata explicitly encourages editors to add claims for which there is no broad consensus, as long as they are corroborated by references. Nevertheless, despite this essential link between content and references, Wikidata's ability to systematically assess and assure the quality of its references remains limited. 
To this end, we carry out a mixed-methods study to determine the relevance, ease of access, and authoritativeness of Wikidata references, at scale and in different languages, using online crowdsourcing, descriptive statistics, and machine learning. Building on previous work of ours, we run a series of microtasks experiments to evaluate a large corpus of references, sampled from Wikidata triples with labels in several languages. We use a consolidated, curated version of the crowdsourced assessments to train several machine learning models to scale up the analysis to the whole of Wikidata. The findings help us ascertain the quality of references in Wikidata, and identify common challenges in defining and capturing the quality of user-generated multilingual structured data on the web. We also discuss ongoing editorial practices, which could encourage the use of higher-quality references in a more immediate way. All data and code used in the study are available on GitHub for feedback and further improvement and deployment by the research community.

\end{abstract}


\maketitle

\section{Introduction}
\label{sec:introduction}

Wikidata is one of the most important sources of structured data on the web. Launched in $2012$, it has grown at an amazing pace, from a little over $4$M entities in early $2013$ to over $80$M in $2020$. It is also one of the main projects of the Wikimedia Foundation, supplying data to a wide range of knowledge-centric activities under a public license \cite{ismayilov2018wikidata,Malyshev2018,thornton2017modeling,chisholm2017learning}. This includes Wikimedia projects, such as Wikipedia itself, where Wikidata provides the data for the summary boxes on the right-hand side of the articles, as well as the inter-language links in the website's sidebar. Additionally, Wikidata entities are linked to entries on multiple digital libraries, such as the National Library of France (BnF) and the Library of Congress (LOC), facilitating integration. Media organisations, such as the Finnish Broadcasting Company, Yle, tag their articles with entity IDs from Wikidata for similar reasons.\footnote{\url{http://wikimedia.fi/2016/04/15/yle-3-wikidata/}} Lastly, big corporations also make extensive use of it in search and question-answering, including Google's Knowledge Graph, or conversational assistants such as Siri and Alexa~\cite{pellissier2016freebase,simonite}.


The more widely used Wikidata becomes, the more critical it is to ensure the quality of its data. As a secondary source~\cite{Vrandecic2012}, Wikidata's reputation is a function of both the adequacy and coverage of its content and of the trustworthiness and relevance of the references that corroborate it. Furthermore, Wikidata openly supports multiple points of view, allowing conflicting claims about an entity to co-exist within the same database, as long as they are backed by references. This design decision has multiple goals, such as committing to knowledge diversity, but also attempting to alleviate the sort of edit-wars~\cite{Vrandecic2013} that tend to occur in Wikipedia~\cite{wikipediawarring}. Wikidata shifts the debate from whether a claim is factually true to whether the sources that back up the claim are relevant and reliable themselves~\cite{Vrandecic2012}. This places even greater emphasis on the quality of the references that are used; they are meant to provide context and supporting arguments for claims to inform discussions and support developers in deciding which information to use.

As Wikidata itself states it \cite{Wikidata6:online}, all claims should, with very few exceptions, have references pointing to specific, authoritative sources that support them. The exceptions are: common-knowledge claims (e.g. the sun is a star); claims where the subject itself is the source (e.g. the Harry Potter book series is written by J. K. Rowling); and claims that specifically refer to an external resource where the verification is straightforward (e.g. the German National Library's ID of the book `Harry Potter and the Philosopher's Stone's' is $4615979$-$4$). The burden to ensure the verifiability of the sources lies with the editor who adds, changes, or restores claims on Wikidata, under the risk of having contributions removed.

In this paper, we explore the quality of Wikidata references in relation to existing policies and recommendations in Wikidata and Wikipedia. We carry out a mixed-methods study, using online crowdsourcing, descriptive statistics, and machine learning (ML), to determine: (i) the relevance; (ii) ease of access; and (iii) authoritativeness of references, at scale and in different languages. Building on previous work of ours \cite{Piscopo2017}, we run a series of microtask crowdsourcing experiments to evaluate a large corpus of references, sampled from triples with labels in several languages. We then consolidate and curate the crowdsourced assessments into a dataset, which we use to train several machine learning models. Given the huge variety of content used as references, we engineer such models to rely on ontological and URL-related features only.

The aim is to create a pipeline to ascertain these three aspects which could be applied for the whole of Wikidata.

The main contribution of this paper is threefold:
\begin{itemize}
    \item Building upon and expanding prior approaches, we devise and evaluate a workflow to perform a large-scale evaluation of references in a knowledge graph. Compared to previous work, the approach we propose is able to take into account references in a wider range of formats, e.g. text, JSON, etc., of quality dimensions, and in multiple languages, therefore being more easily adaptable to real-setting use cases.   
    \item We perform a large scale evaluation of Wikidata references, providing insights about the current state of the project and its evolution since previous analyses, and taking into consideration aspects previously uncovered by the literature. Specifically, we evaluate references along three dimensions, i.e. relevance, authoritativeness, and ease of access, of which the latter has not been previously assessed; furthermore, this is the first attempt to assess the quality of Wikidata references across multiple languages. As noted in~\cite{piscopo2019we}, this is crucial to gauge the success of Wikidata as a multilingual, plural, and diverse source of knowledge.
    \item Based on our quality evaluation, we outline a number of measures aimed at supporting users and improving the quality of references in Wikidata.
\end{itemize}

Overall, Wikidata references are very relevant, fairly to very easy to access, and authoritative. We note some variation though, depending on how the references are encoded and across different languages. We also show that quality assessments can be reliably assisted by machine learning models without content-derived features, encouraging the exploration of such features for further improving assistance and even fully automating the process. The findings help us assess the current quality state of references in Wikidata and identify challenges in defining and capturing the quality of user-generated multilingual structured data on the web. All data and code used in the study are available on GitHub for feedback and further improvement and deployment by the research community.\footnote{\url{https://github.com/gabrielmaia7/wikidata-reference-analysis}}

\paragraph*{Previous publications} In \cite{Piscopo2017} we presented an initial set of crowdsourcing experiments, focusing on references written in English and which had explicitly stated URLs. In this work, we considerably go beyond that study to:
\begin{itemize}
    \item add the ease of access dimension to the analysis;
    \item iterate over the design of the crowdsourcing tasks to support different languages;
    \item increase the sample size, including references in languages other than English, as well as references without an explicit URL;
    \item devise auxiliary methods to obtain URLs based on the claim and reference; and 
    \item develop ML models to tackle our classification tasks, introducing deep learning algorithms that make use of knowledge graph embeddings.
\end{itemize}

From a crowdsourcing point of view, we ask new questions around the feasibility of the task for different languages, and on the quality of the answers. From a machine learning point of view, we propose models that use a wider set of features including ontological, URL-related, and embeddings, but not content-related. More specifically, we want to know if there are patterns in Wikidata that can provide an early indication of whether or not a reference is appropriate to a given claim without having to extract, parse, and interpret the reference, which, given the range of formats and structures, can be a convoluted process that is far from trivial to automate.

The rest of this paper is organized as follows: We start with a background and related work overview in Section \ref{sec:relatedwork}. In Section \ref{sec:methodology}, we explain the research questions we aim at addressing, as well as our crowdsourcing approach to that, describing the microtasks designs used in each experiment. In Section \ref{sec:evaluation} we present the data preparation, experiment designs and their results. We provide a discussion on important aspects of Wikidata in Section \ref{sec:discussion}, under the light of our studies' results. Finally we conclude the paper with a summary of the findings and their implications for future work in Section \ref{sec:conclusion}.
\section{Background and Related Work}
\label{sec:relatedwork}

\subsection{Wikidata}

Wikidata is a knowledge graph, part of the Linked Open Data cloud \cite{Erxleben2014}. Its data are organised as triples (also called \textit{claims} or \textit{statements}) of the form $\langle s,p,o \rangle$, where $s$, the subject, is an \textit{item}, $p$ is a \textit{property}, and $o$, the object, is either an \textit{item} or a \textit{literal}. An item is a Wikidata concept which maps to things in the real world, e.g. physical objects, places, events, concepts, etc. Literals can be either strings, numbers, or timestamps. Properties represent relationships between subjects and objects. Items and properties are identified by URIs starting with either the letter $Q$ or $P$, respectively, and a number. For example, $Q297$ is the item referring to painter \textit{Diego Velázquez} and $P569$ is the property \textit{date of birth}. Claims may also have \textit{qualifiers}, which add specific contextual attributes to the claim itself, and \textit{references}, which provide information about the source for that claim. As of December $2020$, Wikidata has over $80M$ interconnected items and properties, maintained by over $24$ thousands of active users and over $300$ bots~\cite{wikidatastats}. The data is available under a CC0 license~\cite{wikidatalicense}, making it reusable for anyone.

\subsection{References in Wikidata}
\label{sec:references}

\subsubsection{References as nodes in the knowledge graph}
\begin{figure}[ht]
  \centering
  \includegraphics[width=0.7\linewidth]{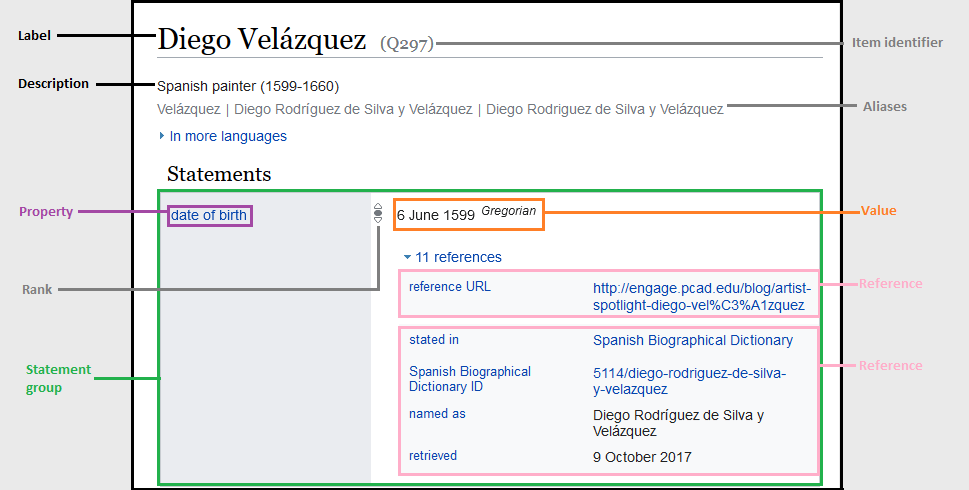}
  \caption{Wikidata's data model as visualised on its website. Entities are also called items, and predicates are also called properties.}
  \label{fig:rdf}
  \vspace*{-12pt}
\end{figure}

\begin{figure}[ht]
  \centering
  \includegraphics[width=0.75\linewidth]{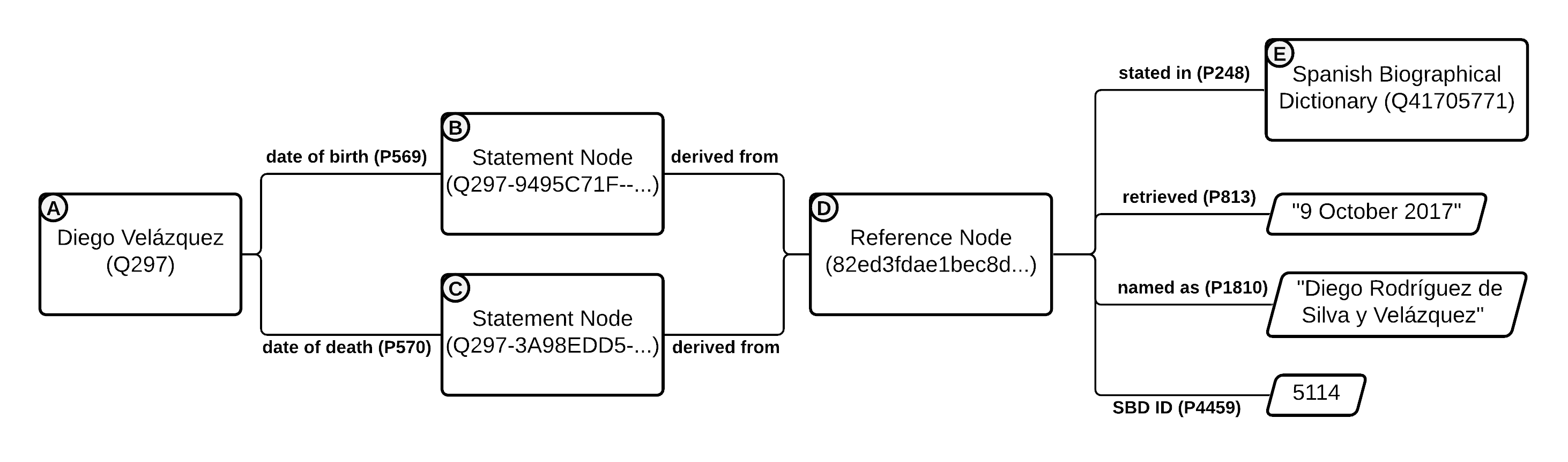}
  \caption{The interconnected nodes which depict Wikidata's information on the birth and death of Diego Velázquez.}
  \label{fig:refnode}
  \vspace*{-12pt}
\end{figure}

Entities, claims, references and literals are rendered in a human-readable format when browsing Wikidata, as seen in Fig.~\ref{fig:rdf}. However, in its underlying data model, Wikidata represents these four concepts by uniquely identified nodes, interlinked by predicates, as seen in Fig.~\ref{fig:refnode}. For clarity, we annotated each node in Fig.~\ref{fig:refnode} with a letter, and non-annotated values are either properties or literals: Nodes $A$ and $E$ are item nodes, representing Velázquez and the Spanish Biographical Dictionary; Nodes $B$ and $C$ are statement nodes and represent the claim of Velázquez's date of birth and death; Node $D$ is a reference node, which is linked to $B$ and $C$. $D$ is also connected to the four pieces of information this reference holds, including node $E$. Thus, references are not attributes of a claim, but rather nodes with their own identifiers, bidirectionally linked to the set of claim nodes which they support. Reference nodes have their own properties, encoding external URLs, access, and retrieval timestamps, pointers to other Wikidata entities, etc. This is shown in Fig.~\ref{fig:refnode}, which depicts a reference node (node $D$) which backs up both the claim about \texttt{Diego Velázquez's ($Q297$)} \texttt{date of birth ($P569$)}, and that of his \texttt{date of death ($P570$)}. Note that statement nodes represent the claims.


\subsubsection{External vs Internal References}

Reference nodes with direct URL to the source are the most intuitive to verify. We analysed them in our previous work~\cite{Piscopo2017}, where we called them \textit{external references}. However, only $60\%$ of references in Wikidata are external references. 

We called the remaining $40\%$ \textit{internal references}, since the URLs needed to access the reference information must be dereferenced within Wikidata. Internal references express provenance in many ways, such as by referencing another Wikidata entity representing the source, by providing an external identifier and a searchable base, or by providing a link to a Wikimedia project page from where the information might have been inferred or imported from.

In this work, we analyse both external and internal references, by exploring methods to find representative URLs for them. We also split the internal references into distinct subgroups, based on how their representative URLs were found, and look at the differences in quality amongst them.

\subsubsection{Wikidata's quality model for references}
\label{sec:qualitymodel}

According to Wikidata's guidelines \cite{Wikidata6:online}, references should be relevant, authoritative, and easy to access. We follow this guidance in our quality assessment; in ~\cite{Piscopo2017} we considered only relevance and authoritativeness. 

To carry this work, we first need to provide our definitions of these quality dimensions. We define as \textit{relevant} those references which support the claims they are linked to by either providing affirmative verbalisation of the claim or by allowing a reader to conclude the claim to be true based on the reference's contents. We define as \textit{authoritative} those references whose sources are considered as reliable sources, according to Wikipedia's definition of reliable sources found in its verifiability guidelines~\cite{wikipediaverif}. In essence, a source is reliable if it is unlikely to intentionally provide false or misleading information. Finally, we define as \textit{easy to access} those references which present the relevant information in a way so that users can find it and understand it with small perceived effort. In this work, we measure the dimensions associated to these definitions, namely relevance, authoritativeness, and ease of access.


When assessing quality along these three dimensions, our work has drawn upon prior work in several areas, which we explore in the remainder of this section: 
\begin{enumerate}
    \item approaches to define and measure the quality of Wikidata (Section~\ref{subsec:wikidataquality}) and Wikipedia (Section~\ref{subsec:wikipediaquality}), not limited to reference quality.
    \item manual (crowdsourced, Section~\ref{subsec:crowdsourcing}) and algorithmic approaches (Section~\ref{subsec:automation}) to ascertain ease of access, relevance and authoritativeness of information on the web, not limited to Wikidata.
\end{enumerate}



\subsection{Quality in Wikidata}
\label{subsec:wikidataquality}

Assessing and improving the quality of linked data and knowledge graphs has been explored by a number of methods~\cite{zaveri2016}, although most focus on the coverage and correctness of the graphs, or on the instance data itself rather than provenance or ontology aspects.
Wikidata has developed several tools for quality assurance, amongst them a website for item constraint violation checks and several bots which run quality and consistency checks.\footnote{\url{https://www.wikidata.org/wiki/Special:ConstraintReport}} 
However, neither of these tools looks at references at all; for references, Wikidata relies exclusively on guidelines on sources~\cite{wikidatasources} and verifiability~\cite{Wikidata6:online}, but enforcement is patchy. We address this gap in our paper -- we essentially propose a supervised machine learning system trained on validated crowdsourced quality judgements to determine whether existing references follow the guidance. 



Farber et al.~\cite{farber2018linked} compare Wikidata's quality with four other knowledge graphs: DBpedia, Freebase, OpenCyc, and YAGO, according to $34$ distinct criteria. They show that overall Wikidata is similar in quality to the other graphs, but achieves better scores in timeliness, schema completeness, and trustworthiness. However, in its assessment of trustworthiness, Farber et al. only check the {\em existence} of references, rather than their quality. Furthermore, they note that the existing guidance on what makes a good reference is subject to interpretation, and point to the fact that some claims may not need references at all. In our work, we consider only the claims with references, regardless of whether or not they are necessary. We also do not measure trustworthiness, but authoritativeness, which is closer to the community guidelines.

Several papers aim to measure the completeness, consistency and accuracy of Wikidata e.g. ~\cite{balaraman2018recoin} for completeness, Piscopo et al.~\cite{piscopo2019we} for a survey on Wikidata, F{\"a}rber et al.~\cite{farber2018linked} for a comparative survey between multiple knowledge graphs. Piscopo et al. note that dimensions such as trustworthiness and accessibility have not yet been adequately covered. The papers that approach accessibility focus on availability and interlinking (e.g.~\cite{farber2018linked,thakkar2016linked}), respectively defined as whether the knowledge graph can correctly answer a given query within a time limit, and whether entities representing the same concept are reachable from one another. These definitions focus on the knowledge graph's data and its inner connections. By contrast, we look at the ease of retrieving information linked to the knowledge graph from sources located outside of it, to check whether the source supports the claim. As such, we use the term \textit{ease of access} instead to avoid confusion between both definitions.


\subsection{Quality in Wikipedia}
\label{subsec:wikipediaquality}

A large body of research~\cite{Zeng2006,adler2008assigning,Hu2007,Dondio2007,Kittur:2008:YET:1460563.1460639,Wohner:2009:AQW:1641309.1641333} has looked at article and reference quality in Wikipedia. We include a summary here because Wikidata and Wikipedia share some commonalities: they are both secondary sources of information, and there is considerable overlap in their communities of editors~\cite{piscopo2017wikidatians}.

Ford et al.~\cite{Ford:2013:GSW:2491055.2491064} have investigated what types of external sources are treated as reliable by Wikipedia contributors. The study found differences in the choice of sources by editors as compared to official guidelines, attesting the prevalence of references to governmental sites, trusted data providers, and popular web sites. This comes at the expense of academic publications, which are Wikipedia's main recommendation of primary and persistent sources. These types of sources are all present in Wikidata as well, and part of our work looks into classifying references in regards to their type of publisher.

Lopes and Carri\c{c}o~\cite{Lopes2008} attempt to measure the accessibility of references and citations in Wikipedia, in terms of providing access to people with disabilities. They show that existing accessibility discrepancies can compromise the overall credibility of Wikipedia. We do not measure accessibility, but ease of access, which we have defined differently, but taking barriers to access into account. 

Machine learning (ML) has been used to measure and improve the quality of Wikipedia's sources. Fetahu et al.~\cite{fetahu2016finding} developed a two-step method: (i) to detect claims in a Wikipedia article needing citations of certain kinds; and (ii) to suggest citations from news sources where they are needed. Anderka et al.~\cite{anderka2012predicting} train a series of $10$ classifiers to discern whether a Wikipedia article suffers from specific quality flaws, like being un-referenced and including primary sources. Our use of ML follows a similar idea; we attempt to model quality dimensions of the references, such as relevance and authoritativeness, based on a crowdsourced annotated dataset.


Although they share many similarities, Wikipedia and Wikidata differ in crucial points. Wikidata gives priority to verifiability rather than to having a consensual point of view~\cite{Vrandecic2012}, thus the primary way to judge the fitness of use of Wikidata information is by assessing whether they contain reliable sources, as opposed to assessing their truthfulness. Also, Wikidata references are structured data, with identifier strings, properties, and values, while Wikipedia's consist of strings only, bearing an URL or not; this allows us to extract features from the structure and feed it to ML models, instead of depending solely on string content and context.

\subsection{Human and Crowd Judgements of Relevance, Authoritativeness and Ease of Access}
\label{subsec:crowdsourcing}

Crowdsourcing is a promising approach for quality assessment for several reasons: 

\begin{enumerate}
    \item it provides access to diverse profiles of web users. Through online crowdsourcing platforms, we reach out to hundreds of thousands of crowd workers from multiple countries~\cite{difallah2018demographics}, who may engage with Wikidata (e.g. as part of Wikipedia's infoboxes), and thus be able to comment on its fitness for use;
    \item it is a well-explored method in current literature, successfully deployed for quality and relevance control in other settings, nearing expert levels of annotation \cite{hossain2015crowdsourcing,ghezzi2018crowdsourcing};
    \item it provides a means to test the quality assessment workflow before proposing it to the Wikidata community and translating it into new guidance and tools.  
\end{enumerate}

\subsubsection{Relevance}
One of our tasks is to determine whether a source is relevant to a Wikidata claim. This is similar to relevance judgements in information retrieval, where people are asked to assess relevance of a document to a user query. Information retrieval routinely uses crowdsourcing for such tasks \cite{muhlberger2014user,Kim2016UsingTC,alonso2009can,blanco2011repeatable}. Similarly, recommender systems ask people for explicit feedback on recommended items \cite{nguyen2020crowdsourcing,Larson2014OverviewOA,Ikemoto2019OnthespotKR}.

We iterated over the relevance task design used in ~\cite{Piscopo2017}, essentially asking the crowd to check if the reference mentions the subject, predicate and object of a Wikidata claim.

\subsubsection{Authoritativeness} 

Crowdsourcing authoritativeness is a more daunting task due to its subjectivity. In \cite{pennycook2019fighting}, crowd workers were given news websites and asked if they knew, and how much they trusted said websites. The authors reported that, although moderated by partisanship and overall familiarity with sources, the crowd was successful in telling mainstream media outlets apart from hyper-partisan and fake news websites, with strong correlation with experts. TRELLIS~\cite{gil2002trusting} is a tool that captures assessments from users about whether they trust information sources in specific contexts. It uses these assessments to provide a score on how authoritative the source is. 

Our approach is different as we do not aim to understand what makes sources, in general, more or less an authority on a subject. For this reason, in our previous work~\cite{Piscopo2017}, we decided not to ask the crowd for their opinions, but to classify the sources based on their publisher (e.g. a government site, an academic paper, a book etc) and author. We used these to decide on authority ourselves, using Wikipedia's own verifiability policy \cite{wikipediaverif}. This approach worked well back then, and we reused it in this paper. In addition, we also modelled publisher and author distributions with machine learning.



\subsubsection{Ease of Access}  
According to Wikidata's policies, web references should be easy to access by at least one person who would like to access them~\cite{wikidataguidelines}. To the best of our knowledge, this is the first study to determine how much effort it takes to consult Wikidata reference.

To inform our crowdsourcing task design, we looked into work done on references and citations in academic literature, web pages, and Wikipedia. Literature is rather broad, tackling anything from availability to their practical usability. Massicotte et al.~\cite{massicotte2017reference} manually analyse web references in electronic theses and dissertations for broken links and content drift. In our approach, we consider these and also include wrong or missing content as possible barriers to access. Liu et al.~\cite{liu2012crowdsourcing} explore ease of access elements closely related to our definition. They crowdsource the analysis of web page design, navigation, interface, information presentation, and more. They conclude that crowdsourcing is a fast and cheap way to identify important usability problems in a website. Like Liu et al., in our study we account for user-centred aspects; we focus on the user's ability to retrieve relevant information from a source while accounting for the page's ease of navigation, as well as for different types of barriers that might prevent the site's retrieval.


\subsection{Automatic Assessment of Relevance, Authoritativeness and Ease of Access}
\label{subsec:automation}

\subsubsection{Relevance} 
In our study, fully automated assessments of relevance are carried only for those reference nodes for which the URL points to an API service and returns a JSON file that follows a consistent format. We check whether the subject, predicate, and object values of a claim are present in the JSON file. In essence, we perform claim verification, where we look in a web resource, namely the JSON file, for evidence.

Automated claim verification is well explored in literature. For instance, 
the FEVER dataset~\cite{thorne2018fever} provides claims in text form, annotated on whether or not they are verified, as well as sentences used as evidence. It also proposes a pipeline for locating and extracting evidence and verifying claims using the evidence. DeFacto~\cite{gerber2015defacto} tackles verification of claims in RDF form, which is much closer to the triple format used in Wikidata. Similar to the pipeline proposed by FEVER, DeFacto locates and extracts content from web pages to use as evidence. Our task is simpler: our claims consist of semantic triples, and our evidence should come from an already determined URL, which gives us a structured JSON file, easier to parse than free text.

\subsubsection{Authoritativeness}
As noted earlier, authoritativeness can be an elusive concept, but there is a large body of literature that tries to link it to observable features of sources. Kleinberg~\cite{kleinberg1999authoritative} propose an algorithm for discovering authoritative web sources on specific topics. They do so through an analysis of links between pages, drawing from graph theory. Our study does not look at the connections of pages but their content and provenance. More recently, Fetahu et al.~\cite{fetahu2016finding} propose a Wikipedia citation discovery algorithm, which evaluates whether a citation comes from an authoritative news source. They do so by defining and extracting features related to authority and feeding them to a random forests classifier. In our work, only a small portion of references are from the news. In addition, our modelling is informed by the verifiability policy in Wikipedia, which is based on the types of sources that are generally held in higher regard by the community. We also take into account Wikipedia's list of deprecated sources, such as publishers of fabricated information, conspiracy theories, and overall unreliable. 

\subsubsection{Ease of Access}
Wu et al.~\cite{Wu2009} analyse availability, an aspect of ease of access, of journal references as URL links, automating the measurement through web browser responses. They provide a classification of unavailability reasons into four major groups: file errors, host name errors, access restrictions, and content update. While we also look at browser responses, we do not classify unavailability reasons at a technical level. Instead, we adopt more descriptive classes, which are more useful for Wikidata editors to understand why a reference needs improvement: technical error, paywall, login request, etc.

For Wikipedia, Tzekou et al.~\cite{tzekou2011quality} look at the rate of dead links on the platform. They analyze $4.5M$ web references across $2M$ articles and account that, as of $2011$, roughly $18.34\%$ of the links were dead, and that around $77.31\%$ of the articles had no dead links in them. However, this only accounts for proper link resolution e.g. no $404$ errors, and fewer than $5$ redirects. While link rot deeply impacts ease of access, our focus also lies on the experience of accessing available links.
\section{Methodology}
\label{sec:methodology}

\subsection{Research Questions}

\subsubsection{Operationalising reference checks with crowdsourcing}
Wikidata's guidelines \cite{Wikidata6:online} prescribe that existing references must be relevant to the claims they support, and that their sources must be authoritative. Furthermore, for a source of information in Wikidata to be considered authoritative, it must also be possible, at least for some users, to feasibly confirm the information present in the source themselves \cite{Wikidata6:online}, meaning that the source should be easy to access within reason. This guidance is aligned with the Wikipedia policy for verifiability \cite{wikipediaverif}. 

Our first aim is to propose manual workflows to operationalise reference checks towards the guidance. We rely on paid microtask crowdsourcing as an affordable approach to collect human judgements at scale \cite{hossain2015crowdsourcing}. The idea is to use paid microtask crowdworkers, via platforms such as Amazon's Mechanical Turk or Prolific, on demand whenever needed to create training data for machine learning algorithms. The algorithms would in time learn to perform such reference quality checks with higher accuracy, removing the need for editors to dedicate precious time and resources to this task. 

Nevertheless, using paid microtasks does not guarantee success; there are common challenges around the feasibility of the task as crowdworkers are unlikely to be familiar with Wikidata and its policy. Thus, we come to the following three research questions:

\newcommand{\RQ}[2]{\begin{itemize}
    \item[\textbf{RQ#1:}] #2
\end{itemize}
}

\RQ{1}{How easy it is to access the sources used in Wikidata?}

\RQ{2}{How authoritative are these sources according to the existing guidance?}

\RQ{3}{How relevant are they for the claims they are associated with?}

We will answer these questions by providing a crowdsourced workflow that outsources answers to these questions to an online set of crowd workers for a representative sample of Wikidata.

As noted earlier, Wikidata has both external and internal references. Adding or changing an external reference is different from an internal one. A crowdsourcing design that works for external references might not work for internal ones. Hence, we are interested to know if this results in variations in relevance, authoritativeness, and ease of access:

\RQ{4}{Are external and internal references of varying quality?}

Further on, we want to look at differences across languages. Wikidata aims to become a source of multi-lingual information, including references in languages other than English where relevant. This is important from a crowdsourcing point of view, as the feasibility of the approach depends on the availability of crowd workers with the relevant language skills:

\RQ{5}{Are references in different languages of varying quality?}

\subsubsection{Automating reference checks with machine learning trained on crowdsourced data}
Paid microtask crowdsourcing is routinely used to create labelled examples for machine learning. Answers from the crowd are checked for quality and aggregated into a dataset, which we then feed into several machine learning models for training, testing, and validation.
It is worth noting that our aim is not to develop a model that checks the content of references for relevance to a Wikidata claim -- this is part of our future work and mentioned in Section \ref{sec:conclusion}. At this stage, we are interested in learning the dependency features of the Wikidata content and surface features of the sources such as type, URL domain etc. This leads us to our final research question:

\RQ{6}{Can we predict relevance, authoritativeness and ease of use of a reference without relying on the content of the references themselves? Can we do this for references in different languages?}

\subsection{Data}
\label{subsec:data}

\subsubsection{Reference Node Extraction}
We have taken a snapshot of Wikidata as of $April 16th, 2020$. At the time, Wikidata consisted of over $82M$ items, from which we extracted a randomized set of $20\%$ (over $15M$ items). From each extracted item, we have parsed all of their claims, alongside the references associated to them. We stored $195M$ unique statement nodes and $13M$ unique reference nodes (we explain in Section \ref{sec:relatedwork} how references are stored in Wikidata as nodes). We used all this information to carry out the analysis described in Section~\ref{subsubsec:stats}, whose results are presented in Section~\ref{subsec:evalstats}.

The corpus includes references in six languages located in different points across the language distributions shown in Section~\ref{subsec:evalstats}. To identify reference languages, we have used a Python port of the language-detection library from \cite{nakatani2010langdetect}, which has over $99\%$ precision on $50$ languages. We chose, in order of coverage in Wikidata: English, Dutch, Swedish, Spanish, Portuguese, and Japanese. Each of these languages corresponds to a large number of references in Wikidata. The least represented of the six, Japanese, which accounts for around $0.19\%$ of references, is estimated to have at least $127$ thousand references. For all six languages, a representative sample size ($95\%$ CI, $5\%$ margin of error) falls between $380$ and $385$ references. Thus, we have sampled $385$ random reference nodes for each of the six target languages, totalling a corpus with $2310$ reference nodes.

For each of the $2310$ sampled reference nodes, we have obtained their id, properties, and a statement node linked to it. The statement node is presented to workers to assess the relevance of the reference. If a reference node was connected to more than one claim, we selected one at random. We also extracted a URL from each reference node, depending on how the reference is encoded, e.g. which properties it has and what other Wikidata objects it connects to. We explain how we did this next.

\subsubsection{URL Extraction}
We identified external URLs by the presence of the \texttt{reference URL (P854)} predicate, which points to an explicit URL. Any reference node lacking that predicate was considered as an internal reference. For those, we obtained URLs via the following series of steps, where if a step finds a URL we kept it, and if it does not we tried the next:

\begin{enumerate}[align=left,style=nextline,leftmargin=*]
    \item If the reference has no explicit URL declared via the \texttt{reference~URL} property, but has an external identifier property, such as \texttt{PUBMED ID (P698)} or \texttt{IMDb ID (P345)}, we obtain the identifier's \texttt{formatter URL (P1630)} and use it to generate an URL. We call these \textit{identifier URLs}.
    \item We check if the reference has the \texttt{Wikimedia import URL (P4656)} property, indicating it has been imported from a Wikimedia page. We then take this page's URL. We refer to these as \textit{wikimedia import URLs}.
    \item We check if the reference has a source declared via the \texttt{stated in (P248)} property. If it does, we look for its official URL or equivalent and use it. We refer to these as \textit{stated URLs}.
    \item We check if the reference has the property \texttt{inferred from (P3452)}, indicating that the information was inferred from another Wikidata item or from elsewhere, and take that item's URL instead. We term these \textit{inferred URLs}.
\end{enumerate}

Based on the over $13M$ unique reference nodes found on our random extraction of $20\%$ of the Wikidata corpus, we project that the entirety of Wikidata has over $66M$ unique reference nodes. To test the aforementioned URL extraction process, and to extract the reference nodes for our crowdsourcing tasks, we drew another random sample of $60K$ unique reference nodes, providing us with a large representative sample ($99.99\%$ CI, $0.25\%$ margin of error) of the Wikidata corpus. We were able to discover a URL for $99.5\%$ of references therein. Out of those, $87\%$ had successful HTTP codes. It is out of these $87\%$ that we have sampled the $2310$ reference nodes which will be shown to the crowd workers.

\subsection{Methods}
\label{sec:methods}
To answer our research questions, we adopted a mix of descriptive statistics, paid microtask crowdsourcing and machine learning. The statistics help us better grasp how references are used by Wikidata editors, what percentage of claims are covered, what predicates are used, and how references are normally coded. We then crowdsource the labelling of a representative sample of reference nodes in six target languages via two different crowdsourcing tasks, one assessing relevance and ease of access, the other authoritativeness. Each reference is allocated to multiple crowd workers. We validate and aggregate the crowdsourced judgements into a dataset using majority voting. Finally, we use the dataset to train different machine learning models.

\begin{figure}[ht]
  \centering
  \includegraphics[width=0.75\linewidth]{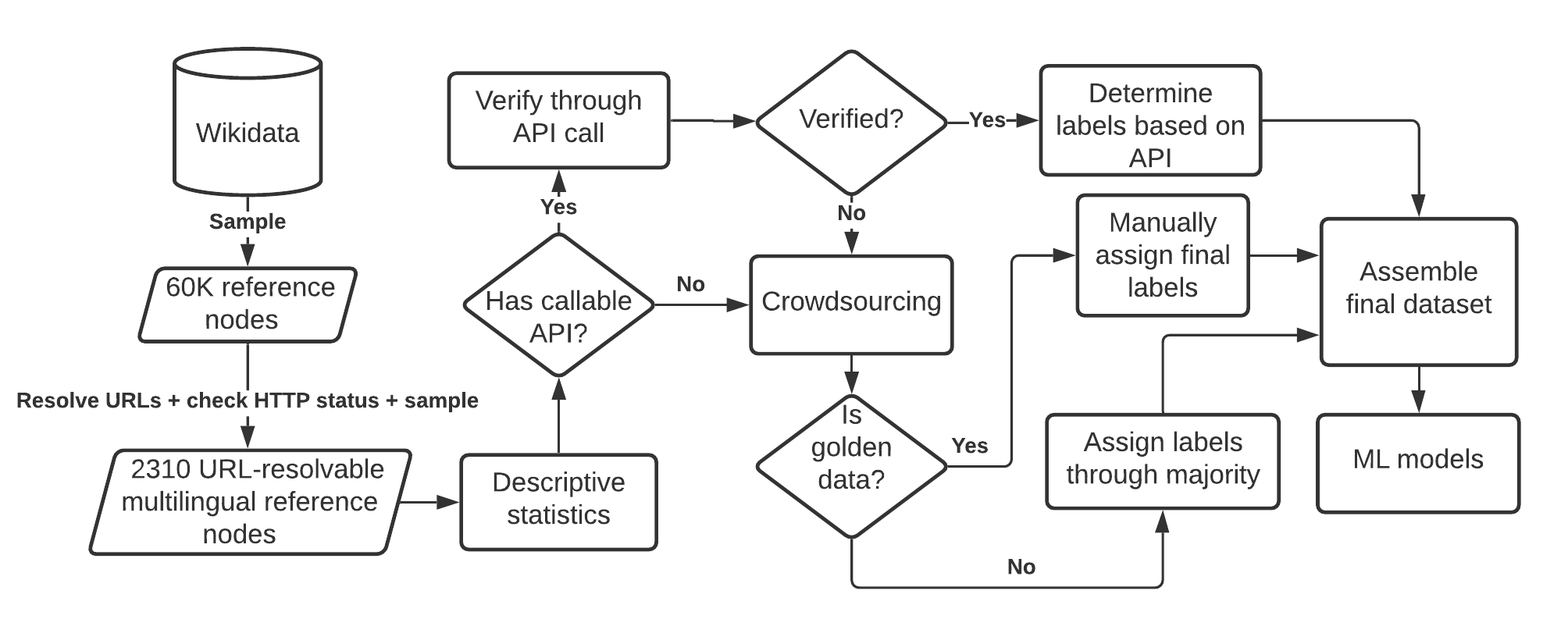}
  \caption{Diagram illustrating the hybrid workflow proposed in this paper.}
  \label{fig:diagram}
  \vspace*{-12pt}
\end{figure}

Fig.~\ref{fig:diagram} illustrates our workflow. We first sample Wikidata for references, calculate some statistics and try to verify them automatically through API calls. Those that could not be automatically verified are given to crowd workers. Some references are manually annotated as golden data, for quality assurance during crowdsourcing. In the end, we gather all annotated samples into a single dataset and train our ML models. In the next sections, we cover each of these steps in more detail.

\subsubsection{Descriptive Statistics}
\label{subsubsec:stats}


To understand the adoption of references in Wikidata, we measured how many claims in Wikidata have at least one reference, defined as the \textit{'reference coverage'}. 

Next, we explored how references are encoded; as explained in Section \ref{sec:relatedwork} editors can select among several predicates to link references to Wikidata claims (e.g. \texttt{stated in}, \texttt{reference URL} and others). The choice of a reference predicate also depends on the availability of relevant information (e.g. retrieval data, curator, etc.), and whether external IDs are available (e.g. PubMed ID, VIAF ID, etc.). In our analysis, we investigated which predicates are used the most how many unique references use certain predicates at least once, and what the most common predicate combinations are. Additionally, since reference predicates have objects, we take a look at the distribution of object types, to understand what type of information references normally encode. In Wikidata, objects of predicates can take one of many types, such as other Wikidata objects, date and time, URLs, external IDs, strings, quantities, etc.

We look further into two reference predicates, \texttt{stated in} and  \texttt{reference URL}, due to their bigger relevance inside the Wikidata topology. To understand what type of provenance the \textit{stated in} predicate usually encodes, we look at what sources are used the most, and aggregate them based on their \texttt{instance of (P31)} predicate objects. To understand the same for the \texttt{reference URL} predicate, we look at the distributions of domain and languages of websites used.

This analysis informed the data pipeline described in Section~\ref{subsec:data}, in particular the extraction of representative URLs for reference nodes. 

The results are presented in Section~\ref{subsec:evalstats}.

\subsubsection{Crowdsourcing}
\label{subsubsec:methcrowd}

In this section, we present a summary of the crowdsourcing task design. Further details are available in Appendix~\ref{appendix:A} for repeatability and reproducibility purposes, including recruitment, monetary rewards, ethics, and extra information on the tasks themselves.

\paragraph{Filtering.}
Before passing the $2310$ references sample to the crowd, we checked how many of those could be verified through simple API calls, without the need of a human in the loop. We did this because we noticed a big portion of representative URLs consisting of API calls that return JSON objects, which then get displayed by the browser. The crowd workers would then only need to search for the correct field, which can be automated.

We wrote scripts to check whether the information given by the Wikidata claim associated to the reference matched the information given by the reference URL, by looking up fields in the JSON objects returned. When a match was not detected, we passed the reference to be crowdsourced alongside the unfiltered ones. When a match was detected, we:
\begin{itemize}
    \item Deemed the reference as relevant;
    \item Determined their author and publisher based on the API service using a purpose-built dictionary (e.g. PubMed being a non-scientific organisation, or Crossref being a non-publishing academic organisation);
    \item Determined their ease of access based on the JSON templates they used at the time of verification, using heuristics.
\end{itemize}

We also chose not to crowdsource the authoritativeness of references pointing to Wikipedia but labelled them as collaborative and self-published by default. We crowdsourced their relevance and ease of access.

The references whose assessment could not be automated in this way were fed to the crowd. 

\paragraph{Crowdsourcing Task Design.}

We designed two crowdsourcing tasks: one for relevance and ease of access, one for authoritativeness. We created versions of each task for the six languages. Both tasks present a set of reference nodes to the workers, alongside some questions the workers need to respond for each reference node. The results can be seen in Section~\ref{subsec:evalcrowdsourcing}.

\paragraph{First Task: Relevance and Ease of Access.}

The first task (\textbf{T1}) collects crowdsourced assessments of relevance and ease of access of references in Wikidata.

Following our definition of relevance in Section~\ref{sec:qualitymodel}, in order to be considered relevant a reference's representative URL should contain, either explicitly or via synonyms, the subject, predicate, and object and link these three in an affirmative sentence. This mirrors Wikidata policy: references are intended as standalone reference; they should not require an argument to be made to support the claim. We capture the ease of access dimension, as defined in Section~\ref{sec:qualitymodel}, in two ways: with an ordinal scale measuring ease of navigation and by looking at the frequencies of a set of access barriers.

Workers are asked to answer three questions for each given reference node:

\begin{itemize}
    \item[ \textbf{(T1.1)}] A \textit{yes} or \textit{no} question on whether the representative URL contains the subject, predicate, and object seen in the given linked claim;
    \item[\textbf{(T1.2)}] A $ 5 $-level Likert scale question on how easy it felt for workers to navigate and find the relevant information on the web page referred to by the URL ($0$ is lowest, $4$ is highest);
    \item[ \textbf{(T1.3)}] A multiple-choice question asking what barriers the worker felt were keeping them from finding the relevant information. We asked this only if the answer to \textbf{T1.1} was negative. 
\end{itemize}



The intuition behind splitting the ease of access assessment into two tasks is due to websites potentially having a combination of information access methods, some more difficult to use than others. Take as an example a Wikipedia page. If the worker is looking for the date of birth of a celebrity, it is often in the infobox to the right side and fairly straightforward to spot. However, if the worker is looking for the celebrity's page in IMDB, that information tends to be found towards the bottom of the page or in related links.




\paragraph{Second Task: Authoritativeness.}

The second task design (\textbf{T2}) aims at collecting crowdsourced assessments on the authoritativeness of references in Wikidata. Our definition of authoritativeness, given in Section~\ref{sec:qualitymodel}, points to Wikipedia's guidelines for reliable sources~\cite{wikipediaverif}, in which it is stated that the authoritativeness of references come from three aspects of the reference: the \textit{type of work} that constitutes the reference, the \textit{author} of the reference, and its \textit{publisher}.

The type of reference can be determined according to the predicates its node possesses. For external references, they would refer directly to web-pages, for which authoritativeness depends on the author and publisher type. For internal references, they may point towards types of work that are by default considered authoritative, such as physical books, scientific publications or legislation documents, or to online versions of those, which would also be considered web-pages. By extracting representative samples from both external and internal references, we can abstract away these differences and treat both types as web pages. 

Web page authoritativeness depends on author and publisher types. To determine them for each reference URL, we follow the approach from our previous work~\cite{Piscopo2017}. We classify author and publisher types, and use them to determine authoritativeness through a look-up table (see Appendix~\ref{appendix:A} for the details). This allows us to avoid asking crowd workers for subjective evaluations of authoritativeness, which is a research topic in its own right~\cite{kkakol2013subjectivity}.

Workers are asked to answer two categorical questions: one on the author type (\textbf{T2.1}), the other on the publisher type (\textbf{T2.2}). When choosing a publisher type that has sub-types, the worker is also asked to select a sub-type (\textbf{T2.3}). 

\paragraph{Quality Assurance}
\label{paragraph:QA}


Paid microtask crowdsourcing does not always work well: the task may be too complex or require specialised knowledge or skills; the design or instructions might be confusing to workers; the prizes might attract workers who don't necessarily have the skills or motivation to do the work well~\cite{xu2015revealing}. We follow established practices in microtask crowdsourcing to alleviate this problem.

Firstly, we employ attention checks, in the form of a language test. At the end of the task's introduction, workers are asked four short and simple "fill-in-the-blank" questions on the reference node's language. Workers are given three tries, and failing them all closes the task. These language questions aim to verify that workers will understand the references, protecting us against spammers and low-quality workers. Also, by inserting a randomized quiz, we give spammers little chance of passing by luck, so they would have to engage, which discourages most of them~\cite{gadiraju2015understanding}, and foil automated scripts.

Secondly, we employ gold-standard checks. Each question in \textbf{T1} and \textbf{T2} consists of six reference nodes, and two among these have gold standard annotations coded in the task. These annotations specify a set of acceptable responses for these reference nodes. If the worker's responses do not align with our gold standard annotations, their contributions are also marked for rejection. 
Gold standards were manually created, and chosen by randomly drawing from the crowdsourcing set of $2310$ reference nodes. To calculate how many annotated references we needed in the gold standard, we first estimated the maximum amount of tasks a worker would likely contribute towards, based on our own experience from similar experiments. Then, we calculated the probability of a worker never encountering a repeated golden set if taking on fewer tasks than the maximum, varying the amount of golden data provided. This probability starts plateauing at $45$ annotations, corresponding to a $90\%$ probability of no repetitions. Thus, for every language set of $385$ references, we put aside $45$ as gold standard.

Lastly, we check the worker's interaction with the task, such as clicks and navigation, and look for suspicious activity. We make use of filters built into Mechanical Turk, recruiting only workers who have cleared more than $100$ tasks, and who have an approval rate of at least $80\%$.

\paragraph{Aggregation and Final Dataset.}
\label{subsubsec:methcrowdaggr}
We distributed our tasks among a large crowd of workers and aggregated the results through majority voting. Most tasks had more than two possible outcomes, and we decided ties by randomly choosing from the most voted options. This yielded more robust measures of our target quality dimensions on each reference node. The robustness scores were obtained by calculating the inter-annotator agreement between the contributing crowd workers~\cite{nowak2010reliable}. We took the scores and compared their values across reference nodes, consisting of external and internal references, as well as between languages, allowing us to tackle RQs $4$ and $5$.

We then merged this crowdsourced dataset with the portion that was automatically checked via API calls. Lastly, the authors manually assigned final annotations to the golden data references and also added them to the crowdsourced dataset. The result was used to train machine learning models.

\subsubsection{Machine Learning\label{subsubsec:methML}}

We obtained our training dataset by joining the manually annotated golden data, the aggregated crowdsourced annotations, and the annotations automated through API calls, yielding 2310 data points. We trained then machine learning classifiers for three tasks:

\begin{enumerate}[align=left,style=nextline,leftmargin=*]
    \item Relevance: Classify between relevant and non-relevant references, using data from \textbf{T1.1}.
    \item Authoritativeness: Classify between authoritative and non-authoritative references, using data from task \textbf{T2}.
    \item Ease of navigation: Assign a level of ease of navigation to the reference, using data from \textbf{T1.2}.
\end{enumerate}

Models for each of the three tasks were trained on slight variations of the same set of features. Such features can be divided into the following four groups, and are further described in Appendix~\ref{appendix:B}.

\begin{itemize}
    \item Features of the representative URL extracted;
    \item Features of the reference node's coding and structure;
    \item Features of the website available through the representative URL;
    \item Features of the claim node associated to this reference node.
\end{itemize}

The variations on the sets of features were due to us testing specific techniques to measure their effect on model performance. Our data has a high amount of categorical features, and we varied the type of encoding used for them: we first tried label encoding on all of them, then tried one-hot encoding on all of them, then tried graph embeddings on ontology-derived categorical features combined with one-hot encoding for the rest. The graph embeddings used were pre-trained on Wikidata dumps by the Facebook Research team at Pythorch-BigGraph~\cite{pbg}.

For one-hot encoding, some features had large numbers of categories, which could raise dimensionality too much. Thus, we joined those least frequent on a single 'other' class, based on a frequency threshold which we varied between 0.1\% and 5\%. We also tested removing features by dropping those with correlation coefficients above a threshold which we varied between 98\% and 90\%. Lastly, we tried the use of PCA as a dimensionality reduction technique. These variations put the number of features between 47 and 716.

Class imbalance was very prevalent, especially for the relevance task. To overcome it, we tested three oversampling algorithms, namely ADASYN~\cite{he2008adasyn}, SMOTE~\cite{chawla2002smote} and BorderlineSMOTE~\cite{han2005borderline}.

Given the properties of our data, we looked for algorithms that are non-linear, handle high dimensionality well, and are robust against imbalanced datasets, outliers, and missing values, as some features were not available for all data points. Thus, we selected the following algorithms: random forest classifier, boosted decision trees, gaussian naive Bayes, support vector machine with an RBF kernel, and neural networks. We tested each of them with the feature variations mentioned earlier. Hyperparameter tuning was also performed, but sensibly to reduce the chances of overfitting. Results from these models, alongside which feature variation and hyperparameters were used, are reported in Section~\ref{subsec:evalml}.
\section{Evaluation}
\label{sec:evaluation}

\subsection{Descriptive Statistics}
\label{subsec:evalstats}

\subsubsection{Reference nodes}

In total, our random sample of $20\%$ of Wikidata contained $195,874,387$ claim nodes, out of which $151,566,485$ had at least one reference, meaning on average $11$ unique claims per reference. In terms of coverage, $77\%$ of Wikidata claims have at least one reference linked to it ($99.9\%$ CI, $0.05\%$ margin of error). While not yet closing on full coverage, this value has been steadily increasing through the years. It is also worth remembering that some claims in Wikidata do not have references at all - for instance, that's the case for claims that can be safely assumed to be common knowledge. A discussion of what is supposed to be referenced and what not in Wikidata is out of the scope of this work, which focuses on the quality of existing references.

The average number of references per claim, coupled with the current coverage, seems to indicate that many references are shared among claims. For example, a reference pointing towards a scientific paper from the VizieR Online Data Catalogue,\footnote{\url{https://vizier.u-strasbg.fr/}}, which publishes astronomical catalogues, is linked to over $1.5M$ claims. Another one identifies imports from the Cebuano-language Wikipedia and is found in over $700k$ claims. Still, most references are used only a few times; $38\%$ of reference nodes link to at most two claims, $52\%$ to at most 7, and $90\%$ to at most 15.

\begin{figure}
\begin{minipage}[c]{0.47\linewidth}
\includegraphics[width=\linewidth]{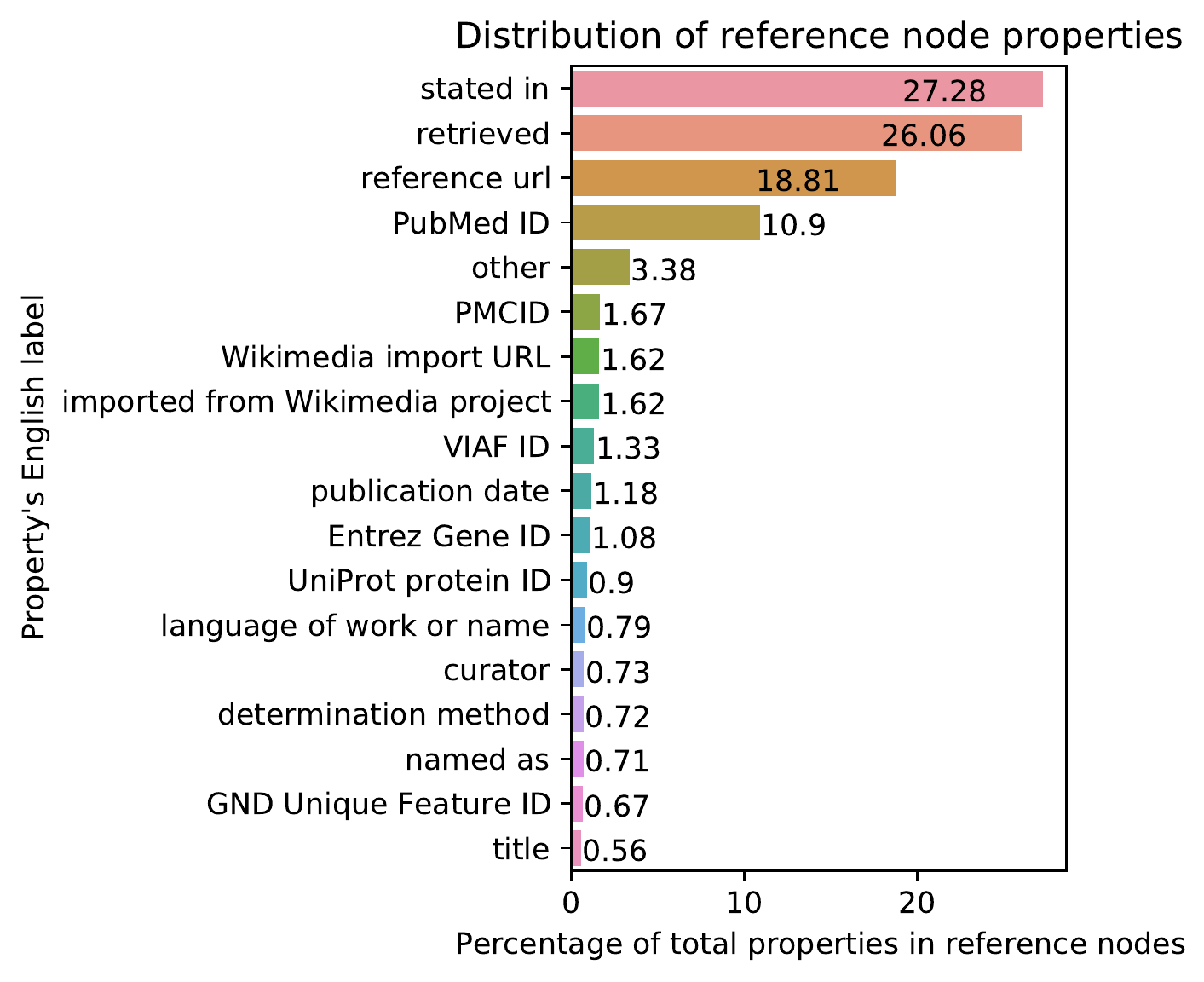}
\caption{The distribution of predicates used by reference nodes to encode provenance information in Wikidata. We aggregate those with less than $0.5\%$ frequency as 'other'.}
\label{fig:refpropstats}
\end{minipage}
\hfill
\begin{minipage}[c]{0.47\linewidth}
\includegraphics[width=\linewidth]{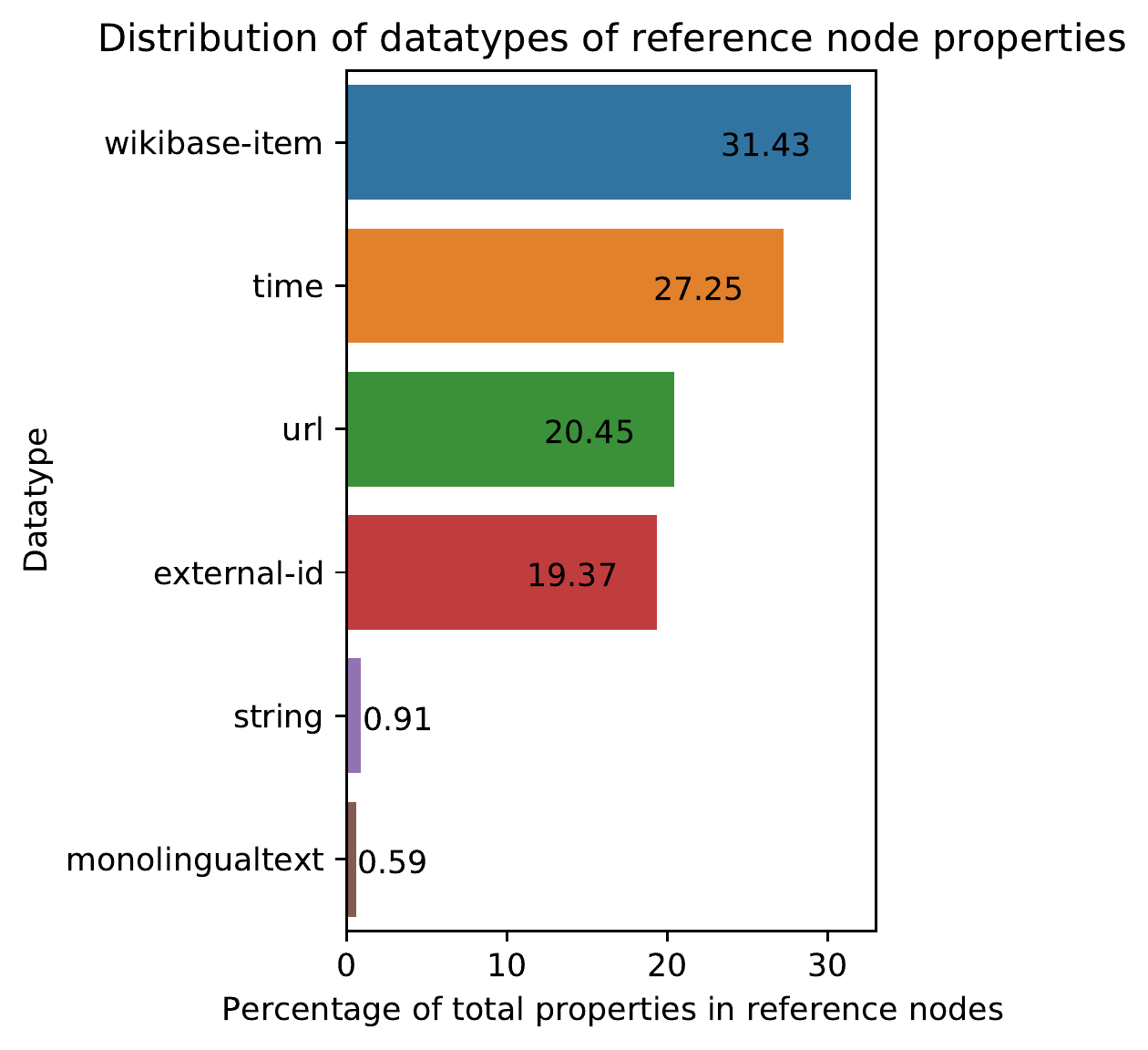}
\caption{The datatypes of objects in reference nodes predicates. The \textit{commonsMedia}, \textit{quantity}, \textit{globe-coordinate}, and \textit{wikibase-property} types had negligible frequency and were excluded.}
  \label{fig:objtypestats}
\end{minipage}%
\end{figure}

In terms of predicates used to connect claims and references, Fig.~\ref{fig:refpropstats} shows that \texttt{reference URL} and \texttt{stated in} make up almost half of the sample analysed, and \texttt{retrieved (P854)} is at around a quarter. The remaining quarter of the sample uses mostly external ID predicates and miscellaneous, and a smaller fraction ($3.24\%$) refers to Wikimedia imports.

We then look deeper at the coverage of predicates \texttt{stated in}, \texttt{reference URL}, \texttt{Wikimedia import URL}, and \texttt{imported from Wikimedia project}, which communicate provenance. About $86\%$ of reference nodes have a \texttt{stated in} predicate, linking to a source with a Wikidata object representation, and about $60\%$ of reference nodes point to a web page, directly linked via the \texttt{reference URL} predicate. Coverage for the two Wikimedia import predicates is at about $5\%$ for each. This means that the majority of references in Wikidata can be classified as \textit{external} - as discussed in Section~\ref{sec:references}, these are references that have a direct URL to an outside source. However, the most common predicate for provenance coding is still \texttt{stated in} predicates, which point to sources represented by Wikidata entities, internally. As reference nodes may contain multiple predicates, many references are introduced using both \texttt{reference URL} and \texttt{stated in}. On the other hand, \texttt{Wikimedia import URL} and \texttt{imported from Wikimedia project} are mostly used on their own or in combination with each other, but not with other predicates ($93\%$ of the references with import URL and $92\%$ of references with imported from project, respectively).


Reference predicate objects can have one of many types (Fig.~\ref{fig:objtypestats}). This confirms that most reference predicates are used to link claims to Wikidata objects. It also shows that external IDs, when aggregated, are used almost as much as direct URLs.

\subsubsection{Stated in}

\begin{figure}
\begin{minipage}[c]{0.47\linewidth}
\includegraphics[width=\linewidth]{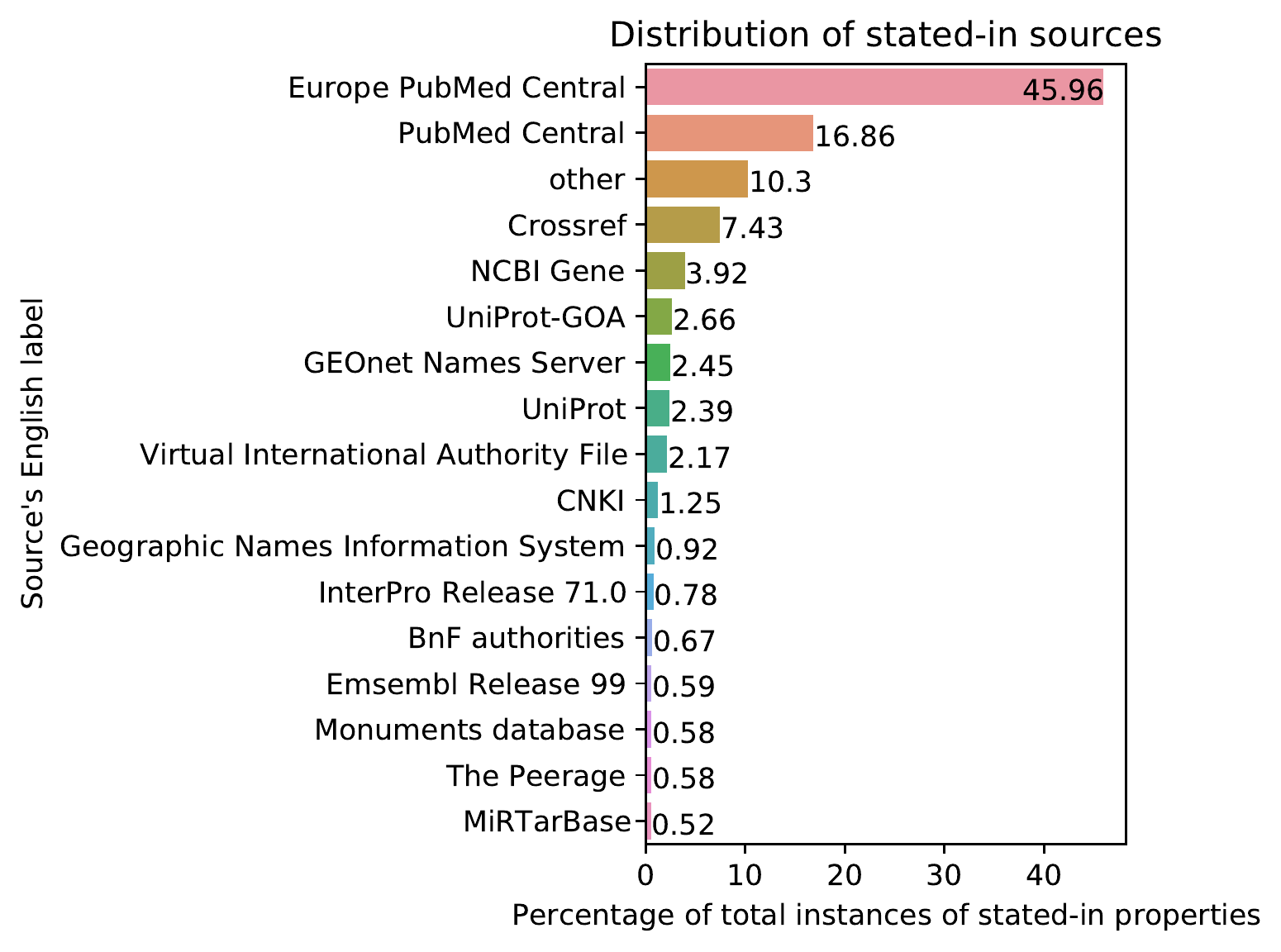}
\caption{The distribution of Wikidata entities used as objects of \texttt{stated in} predicates. We aggregate those with less than $0.5\%$ frequency as 'other'.}
\label{fig:statedindistwikidata}
\end{minipage}
\hfill
\begin{minipage}[c]{0.47\linewidth}
\includegraphics[width=\linewidth]{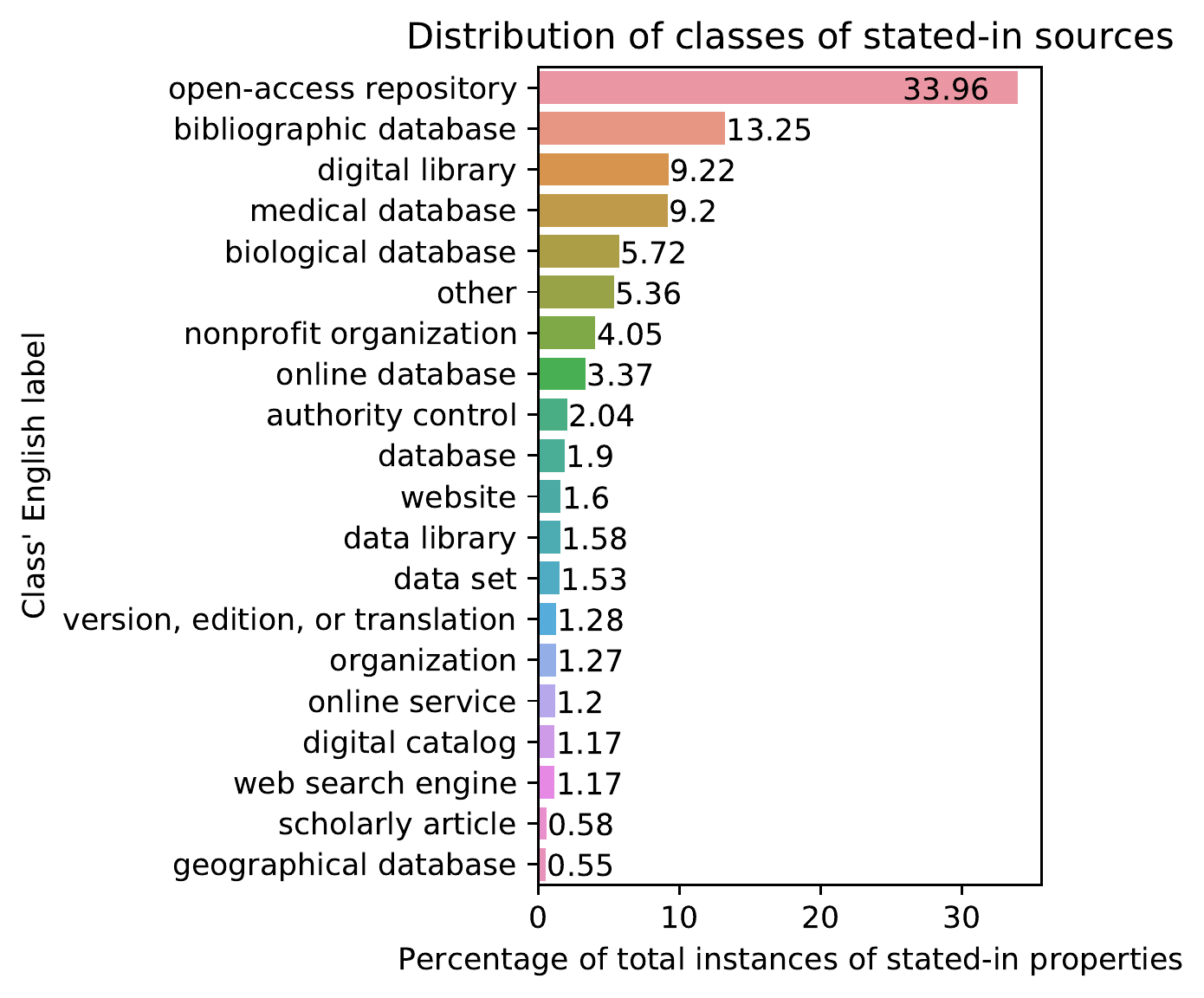}
\caption{The distribution of classes of Wikidata entities used as \texttt{stated in} objects. We aggregate those with less than $0.5\%$ frequency as 'other'.}
  \label{fig:statedinclassdistwikidata}
\end{minipage}%
\vspace*{-12pt}
\end{figure}

Fig.~\ref{fig:statedindistwikidata} shows the distribution of Wikidata entities used as objects of \texttt{stated in} predicates in the reference nodes. We note that most \texttt{stated in} sources are related to scientific publications, with over half of all pointing towards PubMed Central and Europe PubMed Central, which are archives of life sciences journal literature, and Crossref, which deals with information on scientific publications. After that, there are biology bases, such as NCBI Gene and UniProt. 

Wikidata has a predicate for encoding class-entity relations: \texttt{instance of (P31)}. The entity \texttt{PubMed Central (Q229883)}, for instance, is the subject of a \texttt{instance of} predicate where the object is the entity \texttt{open-access repository (Q7096323)}. For each Wikidata entity acting as objects of \texttt{stated in} predicates, we looked at their \texttt{instance of} predicates in search of their classes. The final distribution of classes can be seen in Fig.~\ref{fig:statedinclassdistwikidata}. Over a third of \texttt{stated in} sources are open-access repositories, such as PubMed and Zenodo. It is important to notice here how the great majority of sources are either virtual or virtually accessible, such as databases, data repositories, datasets, digital catalogues, etc. Only a small portion, such as organisations and version, edition or translation, might not have a representative URL. 

\subsubsection{Reference URL}

The URLs used as objects of the \texttt{reference URL} predicate might be repeated between different reference nodes. For example, a URL linking to a downloadable zip file in the \url{geonames.usgs.gov} website, with information on $2016$ US Federal Codes, is shared by $6.3$k unique reference nodes, the highest figure of its kind in our sample. At the time of writing, it returned a $404$ code; in fact, from the top $5$ most shared reference URLs, $3$ gave errors. This suggests that the platform needs to implement more effective policies to check for rotten links in its references.

\begin{figure}
\begin{minipage}[c]{0.47\linewidth}
\includegraphics[width=\linewidth]{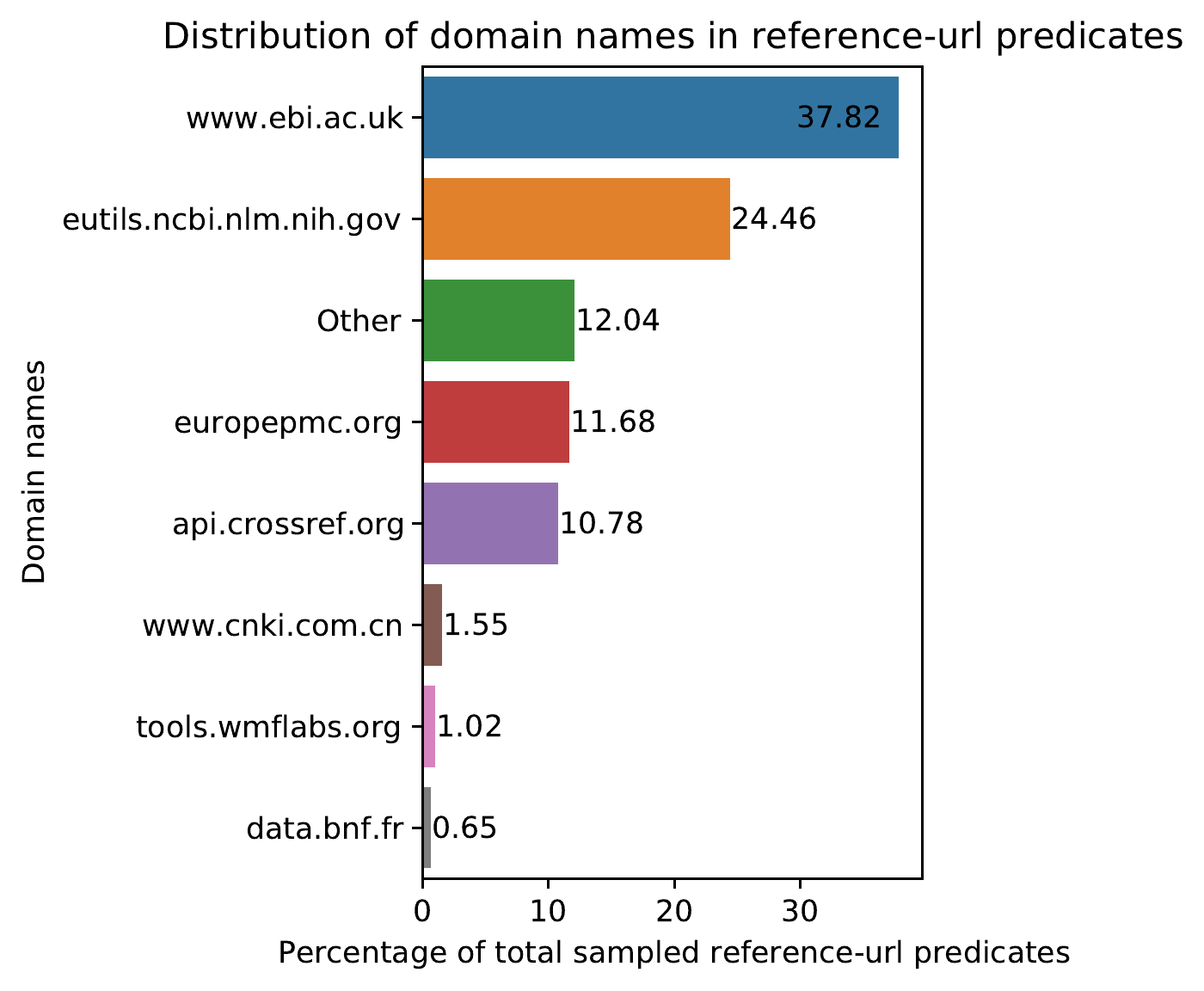}
\caption{The distribution of URL domain names found in \texttt{reference URL} predicates of reference nodes. All domains with less than $0.5\%$ frequency were aggregated under \textit{other}.}
\label{fig:domains}
\end{minipage}
\hfill
\begin{minipage}[c]{0.47\linewidth}
\includegraphics[width=\linewidth]{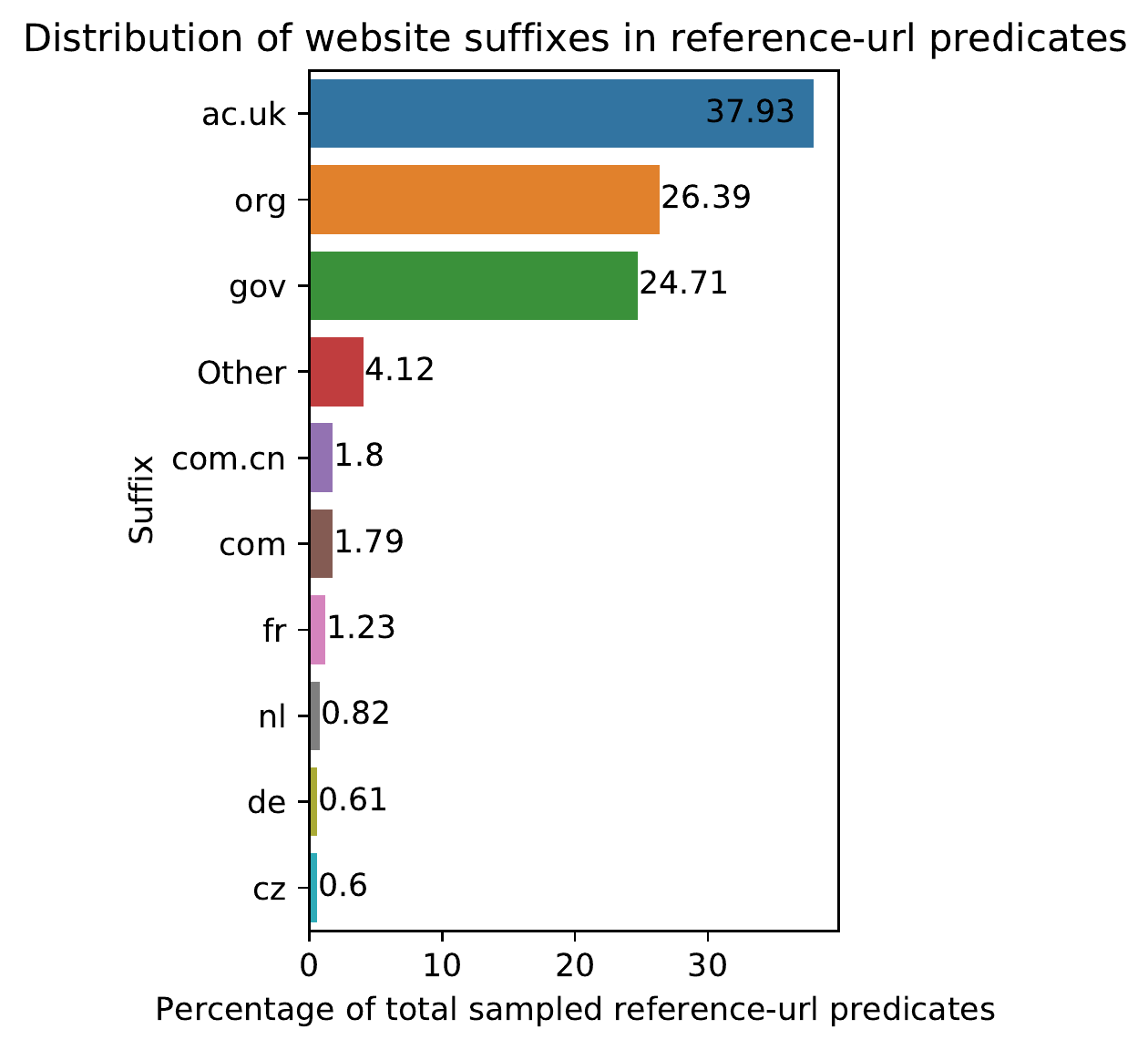}
\caption{The distribution of URL suffixes found in \texttt{reference URL} predicates of reference nodes. All suffixes with less than $0.5\%$ frequency were aggregated under \textit{other}.}
  \label{fig:suffixes}
\end{minipage}%
\end{figure}

We found $45k$ unique domain names in the URLs in the sample. Fig.~\ref{fig:domains} shows the distribution of domain names. Over a third ($37.85\%$) of \texttt{reference URL} predicates point towards \url{ebi.ac.uk}, a website for biology data. After it, we have NCBI (National Center of Biotechnology Information), PubMed Central Europe, and Crossref. The vast majority of \texttt{reference URL} predicates thus point towards biology data or scientific publication data. This matches the use of \texttt{stated in} predicates. As these two predicates plus \texttt{PubMed Id (P698)} and \texttt{PMCID (P932)} add up to almost $60$\% of reference predicates, this reinforces the previous comment on the prevalence of scientific information, in particular biology in the knowledge graph.

Fig.~\ref{fig:suffixes} further shows that the vast majority of \texttt{reference URL} links are hosted by academic (ac.uk), organisational (org) or governmental (gov) entities, while also hinting that the majority of them are written in English. To confirm this, we sampled $18$k objects from \texttt{reference URL} predicates ($99\%$ CI, $1\%$ margin of error), and parsed their contents using a re-implementation of Google's language detection library~\cite{nakatani2010langdetect}. Firstly, $92\%$ of them returned a status code of $200$, indicating that references from this predicate do not suffer greatly from link rot. The results of the language distribution are shown in Fig.~\ref{fig:languages}, confirming that English is the most dominant language. For our crowdsourcing experiments later on, we chose to pick languages located in different points of this distribution: English, which represents the majority; Dutch, and Swedish, relatively well represented when compared to the others; Spanish and Portuguese, in the middle of the distribution; and Japanese, near the trailing end.

The type of content returned by references from this sample is mainly JSON ($67.78\%$), followed by HTML ($31.53\%$), with other types present in very small numbers. This means these \texttt{reference URL} predicates contain more structured data, which is, compared to free text, aimed mostly at machines rather than people. While people can read JSON, the distribution does raise questions about how relevance checking is supposed to work in Wikidata - in this work, we use a manual process (crowdsourced) to generate labels to train a machine learning model. For future studies, it would be interesting to get a more thorough understanding of how people go about checking for relevance when the reference is using formats meant for machine consumption.

\subsubsection{Representative URL extraction}

\begin{figure}
\begin{minipage}[c][7cm]{0.47\linewidth}
\includegraphics[height=5cm]{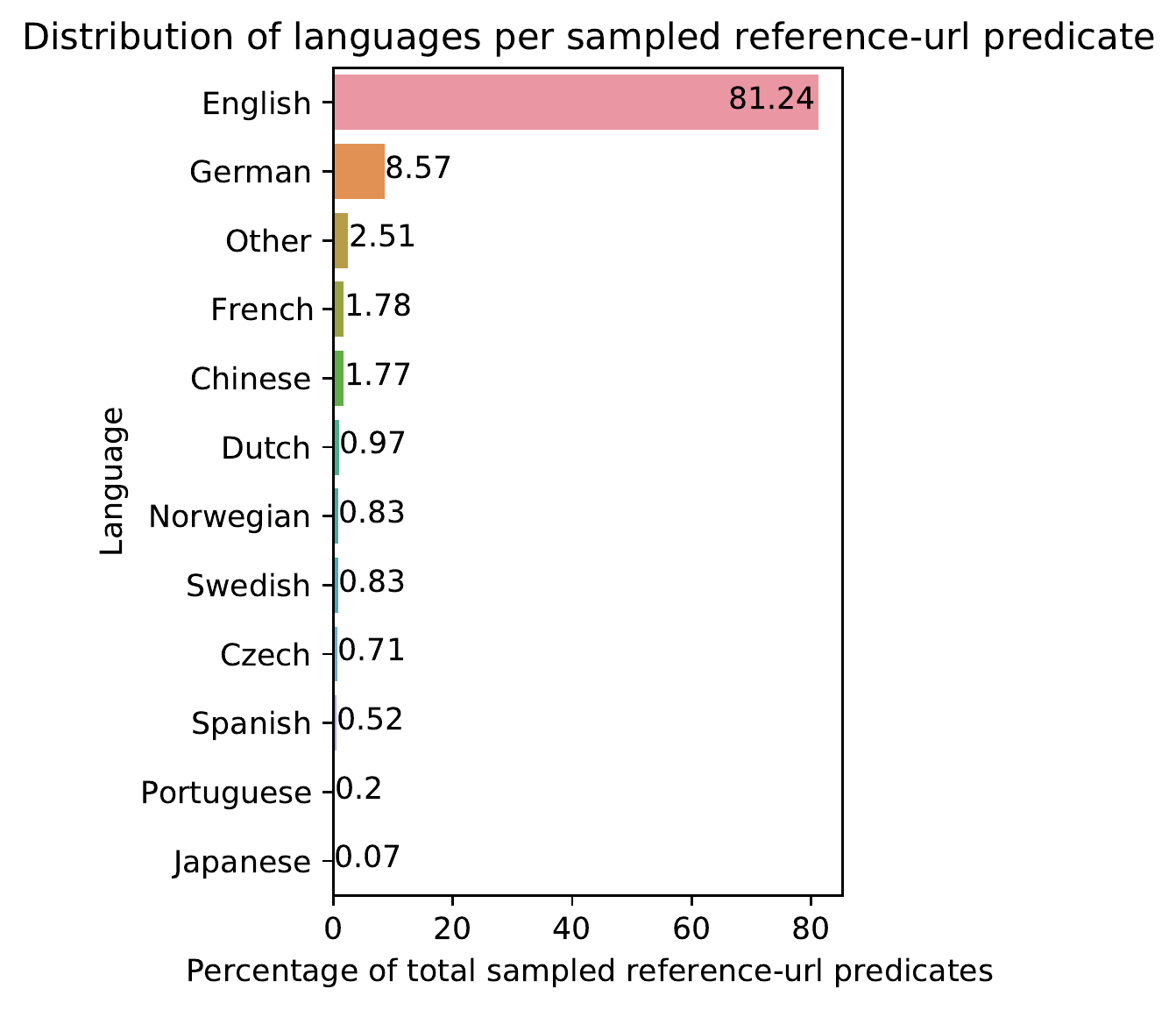}
\caption{The distribution of languages in a sample of the \texttt{reference URL} predicates found in reference nodes. All languages with less than $0.5\%$ frequency were aggregated under \textit{other}, except for those targeted by our study.}
\label{fig:languages}
\end{minipage}
\hfill
\begin{minipage}[c][7cm]{0.47\linewidth}
\includegraphics[height=5cm]{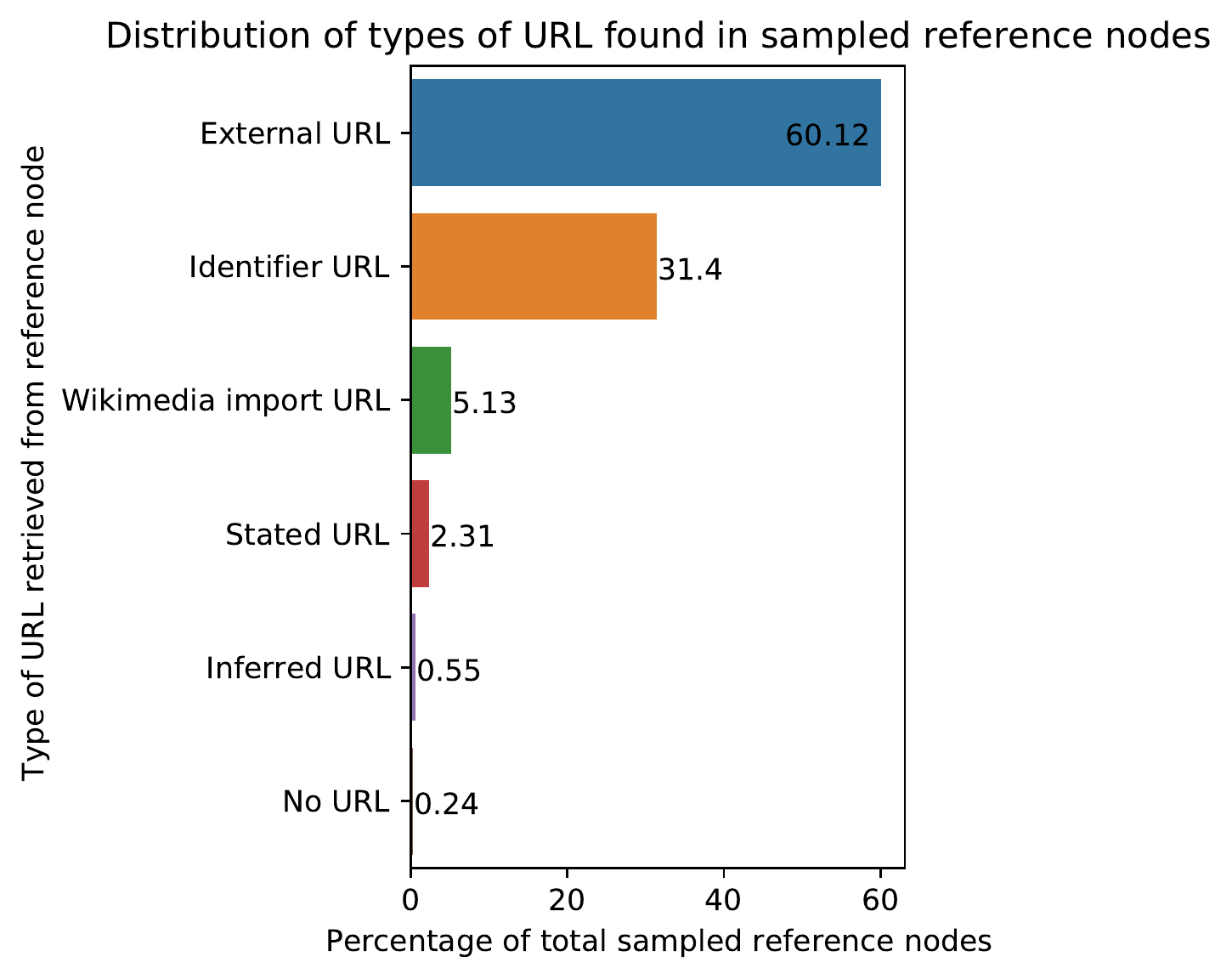}
\caption{The distribution of URL types in a sample of the \texttt{reference URL} predicates found in reference nodes.}
  \label{fig:urltypes}
\end{minipage}%
\vspace*{-9pt}
\end{figure}

After applying our representative URL extraction to a sample of $60k$ reference nodes, as described in Section~\ref{subsec:data}, we observed the distribution of URL types seen in Fig.~\ref{fig:urltypes}. Over $90\%$ of reference nodes either have a direct URL pointing at some web resource, or an external identifier linking it to some digital catalogue or base. These, plus Wikimedia import URL, are the cases where a link to the specific page, work or text where the claim is made can be coded. In the other URL types, namely stated URL and inferred URL, the reference nodes point only towards other Wikidata objects, making it difficult in most cases to identify where the claim came from. For example, the object might be the English Wikipedia, a book, or another database, but it is challenging to know in more detail. This lets us conclude that, for over $90\%$ of reference nodes with representative URLs extracted, these can potentially be accessed and the claim can be potentially inspected by a human. We test and confirm that this is indeed the case through the use of crowdsourcing in our study.

\subsection{Crowdsourcing}
\label{subsec:evalcrowdsourcing}

\subsubsection{Automated Checks on References in Different Languages}
As mentioned in Section \ref{sec:methods}, the reference nodes whose representative URLs could be checked by API calls were automatically evaluated. This number was higher in some languages than others: $71.17\%$ for English, $66.7\%$ for Swedish, $38.5\%$ for Japanese, $28.52\%$ for Portuguese, $25.88\%$ for Spanish, and $8.23\%$ for Dutch. 

There were also differences in the content covered by each language sub-sample. English references, as we have seen so far, have the bulk of their data ($78\%$) associated to biology and scientific publications, whose URLs point to services with APIs, such as Crossref, EBI, NCBI, and PubMed, which all return JSON documents we can parse and check with a script. Swedish references mostly ($70\%$) tend to refer to the Wikimedia foundation heritage monument archives and the National Library of Sweden's LIBRIS service, both of which have APIs and are automatable as well. The remaining $30\%$ link either to Wikipedia or museums, mainly.

References for Japanese are $41\%$ to Wikipedia, and $39\%$ to the Virtual International Authority File (VIAF); the latter can be checked automatically. The remaining $20\%$ point mainly to Yahoo, baseball, and government websites. A similar pattern is seen in Portuguese, where the two largest groups of references are $28\%$ pointing to Wikipedia, and $13\%$ to VIAF. For Spanish, $33\%$ are Wikipedia references, and the automatable portion of them links to biology/scientific databases just like the English ones. Anecdotally, the Spanish Wikipedia references seem to focus on pages about football players.

Dutch is by far the least automatable part of the corpus, mainly because the majority ($81\%$) of its references linking to institutes of art, museums, history, and heritage centres, which mostly do not offer APIs. When such APIs are available, they do not have the same information linked in the web pages. There are some references pointing to EBI or VIAF, but these are much less prominent than for the other languages.

We see thus that each language has a niche, such as biology for English and culture for Dutch. Also, while referencing Wikipedia is something that was very frequent in the earlier days of Wikidata, the knowledge graph has come a long way to replace references to Wikimedia projects with more authoritative sources. However, this has not yet reached all languages - $70\%$ of Swedish references points to a Wikimedia tool, while big portions of Japanese, Portuguese, and Spanish references are Wikipedia references.

The remaining references in the sample, which could be checked automatically, were annotated by the crowd.

\subsubsection{Individual Annotations}

We collected $15.3$K individual annotations from crowd workers on Amazon Mechanical Turk. The majority of these have come from workers who performed the task six times or less, but we also had some workers who contributed much more. The top ten workers in number of contributions range from $59$ to $272$ individual assignments.

Per language, in descending order, we had $4620$ annotations for Dutch, $2910$ for both Spanish and Portuguese, $2130$ for Japanese, $1410$ for English and $1320$ for Swedish. The difference is due to the fact that for each language, we could check a different share of the sample automatically, leaving us with sub-samples of different sizes to be crowdsourced. As per microtask type, we had $9270$ annotations coming from \textbf{T1} and $6030$ for \textbf{T2}. We were able to describe author and publisher types of references pointing to Wikipedia automatically, hence we removed them from \textbf{T2}.

\begin{figure}[ht]
  \centering
  \includegraphics[width=7cm]{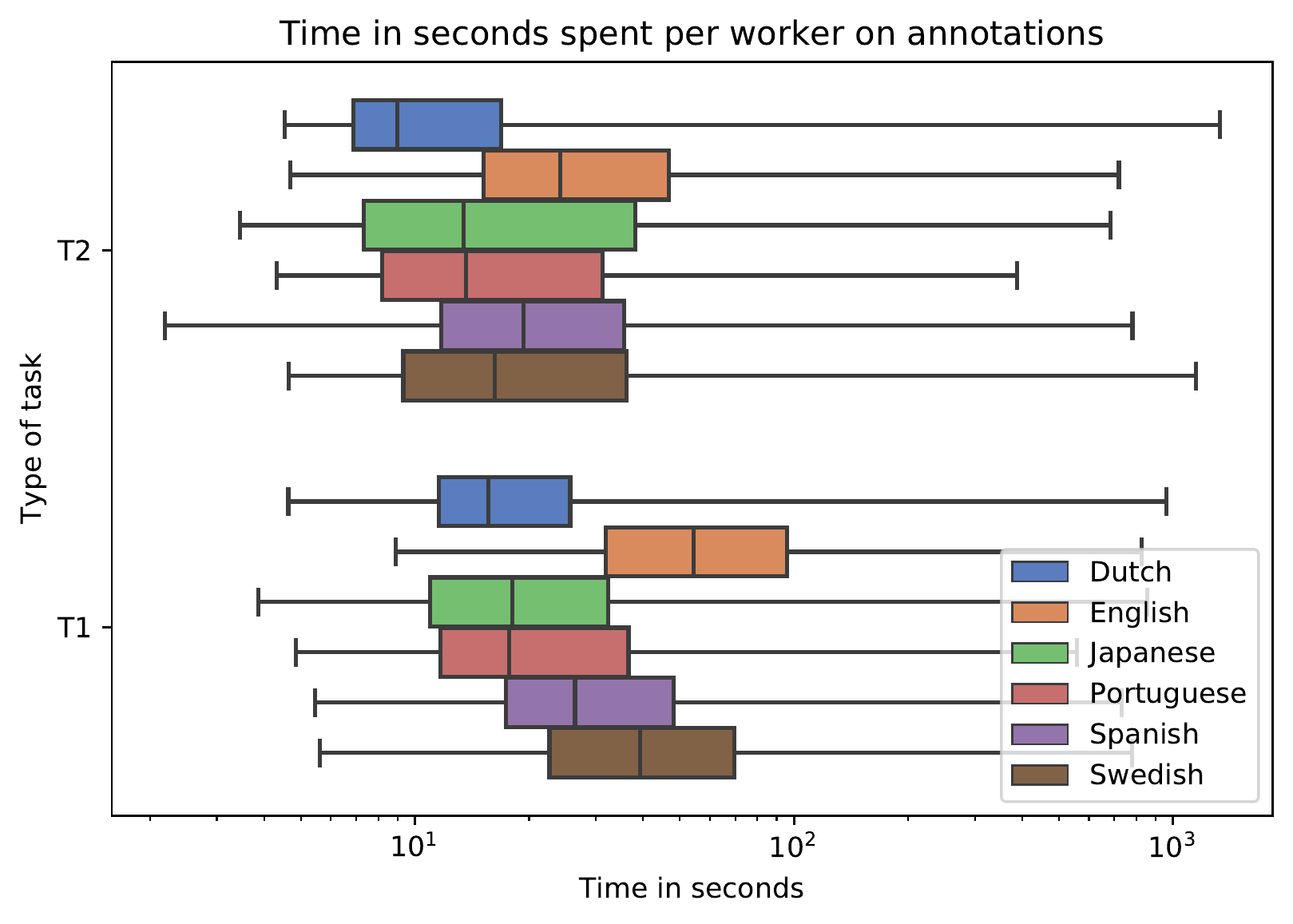}
  \vspace*{-10pt}
  \caption{Time in seconds spent by workers per task type and language group.}
  \label{fig:timespent}
\vspace*{-12pt}
\end{figure}

Workers spent more time in \textbf{T1} than \textbf{T2} (Fig.~\ref{fig:timespent}) as the former has a higher cognitive load. English was the language group that required the most time, and we suspect that it is due to references in this group being more difficult to understand, as its majority refers to either biology or other scientific data. We discuss this in more detail in Section~\ref{sec:aggreg}. The other language groups have time distributions which we suspect vary based on the prevalent type of reference found in them (e.g. museums, government data, libraries, etc). The mean time spent by workers also validates the monetary reward given, which is explained further in Appendix~\ref{appendix:A}. 

\subsubsection{Quality of Annotations}

We used the following set of metrics to measure inter-annotator agreement: Krippendorff's alpha~\cite{krippendorff}, as it works with nominal, ordinal, and interval data by employing distinct difference functions, and is widely adopted in literature; and Fleiss' kappa~\cite{fleiss1971measuring} and Randolph's kappa~\cite{randolph2005free} for categorical annotations. We used the latter to avoid the high agreement, low kappa paradox~\cite{feinstein1990high,randolph2005free} Fleiss' kappa is known to be prone to when the true class distribution of the data is imbalanced. This is true to our data, as seen in Section~\ref{sec:aggreg}.
We hence use Randolph's kappa as well, as it does not have the same issue as Fleiss' kappa; it is margin-free and makes no assumption as to the distribution of classes \cite{randolph2005free}. 

The agreement levels are listed in Tables~\ref{tab:agreement} and \ref{tab:agreement_url}. As the question on ease of navigation asks for interval answers, we can not use kappa measures with it. Instead, we adopted the interpretations recommended by Landis \& Koch \cite{landis1977measurement}, in which agreements lower than $0$ are poor, $0$ to $0.2$ are slight, $0.2$ to $0.4$ are fair, $0.4$ to $0.6$ are moderate, $0.6$ to $0.8$ are substantial, and $0.8$ to $1$ are almost perfect. By looking at Krippendorff's alpha (K-alpha), we note that all questions, grouping all languages together, have moderate to substantial agreement, with $3$ out of $6$ questions showing substantial agreement. Randolph's kappa (R-kappa) shows almost all questions to have substantial agreement, except \textbf{T1.3} with $0.592$ which comes close to having it.

Fleiss' kappa (F-kappa) shows interesting results, especially when compared to the other two metrics. In \textbf{T1.3}, it reaches negative values in three languages. R-kappa and K-alpha seem to disagree, which might be a case of the high agreement, low kappa paradox discussed earlier. 

Looking at the individual languages, we can see some trends. First, English shows the lowest agreement scores overall, often by a wide margin. In \textbf{T1.1} and \textbf{T1.2}, we believe this to be due to references in English being harder to understand, due to their type of content, and thus their relevance harder to state, and workers with different skills needing different forms of interaction with the websites to extract the necessary information. One example is on gene and protein data, where whether a specific gene encodes a specific protein might be clear to some by reading its name, but others needing to read the whole page to try and understand this concept.

Second, \textbf{T1.3} seems to have lower agreements overall. This is expected, as different workers might have different elements keeping them from accessing information. For example, some workers used company notebooks, which blocked them from accessing certain websites due to security issues.

There are some differences with respect to URL type. We have aggregated all non-external references as being internal (see Section \ref{sec:references} and \ref{sec:methods}). External references seem to draw much more agreement than internal references, except for question \textbf{T2.1}. This shows that, for every question except \textbf{T2.1}, having direct access to the information intended to be seen makes it easier to understand it, in contrast with using the predicates of the reference node to infer an URL. As for \textbf{T2.1}, this is a curious result that should be further explored, but does not deny the overall trend of external references drawing better agreement levels.

\begin{table}[h]
\centering
{\small
\setlength{\tabcolsep}{3pt}
\renewcommand{\arraystretch}{0.9}
\vspace*{0pt}
\begin{tabular}{|c|l|l|l||l||l|l|l|c|}
\hline
Question  & \parbox{1.1cm}{F-Kappa}      & \parbox{1.15cm}{R-Kappa}      & \parbox{1.1cm}{K-Alpha}     & Language   & \parbox{1.1cm}{F-Kappa}   & \parbox{1.15cm}{R-Kappa} & \parbox{1.1cm}{K-Alpha} & Question  \\ \hline
T1.1    & 0.507702    & \textbf{0.745626}      & 0.537647         & All        & 0.496904 & \textbf{0.781481} & \textbf{0.645795}     & T2.1     \\ \cline{2-8}  
        & \textbf{0.627820}    & \textbf{0.804444}      & \textbf{0.660418}         & Dutch      & 0.378613 & \textbf{0.950000} & \textbf{0.604762}     &          \\ \cline{2-8}  
        & 0.328039    & 0.438532      & 0.403235         & English    & 0.486781 & \textbf{0.665993} & 0.595334     &          \\ \cline{2-8}  
        & 0.583579    & \textbf{0.828837}      & 0.589590         & Japanese   & 0.563154 & \textbf{0.736626} & \textbf{0.701034}     &          \\ \cline{2-8}  
        & 0.459625    & \textbf{0.740891}      & 0.483134         & Portuguese & 0.335173 & \textbf{0.767544} & \textbf{0.675508}     &          \\ \cline{2-8}  
        & 0.440176    & \textbf{0.766154}      & 0.467675         & Spanish    & 0.463310 & 0.586147 & 0.557485     &          \\ \cline{2-8}  
        & 0.464771    & \textbf{0.687805}      & 0.493689         & Swedish    & 0.349380 & \textbf{0.727928} & 0.469866     &          \\ \hline\hline
T1.2    & NA          & NA            & \textbf{0.614661}         & All        & \textbf{0.603499} & \textbf{0.695833} & \textbf{0.657171}     & T2.2     \\ \cline{2-8}  
        & NA          & NA            & \textbf{0.693767}         & Dutch      & 0.543380 & \textbf{0.816053} & \textbf{0.612898}     &          \\ \cline{2-8}  
        & NA          & NA            & 0.418053         & English    & 0.425262 & 0.496970 & 0.477276     &          \\ \cline{2-8}  
        & NA          & NA            & \textbf{0.654111}         & Japanese   & 0.514036 & 0.583704 & 0.578147     &          \\ \cline{2-8}  
        & NA          & NA            & 0.569897         & Portuguese & \textbf{0.657978} & \textbf{0.714211} & \textbf{0.716573}     &          \\ \cline{2-8}  
        & NA          & NA            & \textbf{0.602839}         & Spanish    & 0.545931 & 0.594026 & 0.572834     &          \\ \cline{2-8}  
        & NA          & NA            & 0.591861         & Swedish    & 0.392275 & \textbf{0.755068} & 0.510743     &          \\ \hline\hline
T1.3    & 0.405330    & 0.592810      & 0.415045         & All        & 0.515551 & \textbf{0.604419} & 0.572782     & T2.3     \\ \cline{2-8}  
        & 0.393451    & \textbf{0.606061}      & 0.448659         & Dutch      & 0.443853 & \textbf{0.743421} & 0.503451     &          \\ \cline{2-8}  
        & -0.184211   & 0.280000      & 0.277645         & English    & 0.282827 & 0.339945 & 0.340493     &          \\ \cline{2-8}  
        & -0.086957   & \textbf{0.786667}      & \textbf{0.637805}         & Japanese   & \textbf{0.755068} & 0.417010 & 0.438724     &          \\ \cline{2-8}  
        & -0.054560   & 0.383333      & 0.249919         & Portuguese & \textbf{0.610643} & \textbf{0.679187} & \textbf{0.674295}     &          \\ \cline{2-8}  
        & 0.518600    & \textbf{0.612121}      & 0.542760         & Spanish    & 0.511959 & 0.574262 & 0.536547     &          \\ \cline{2-8}  
        & 0.087838    & 0.200000      & 0.288190         & Swedish    & 0.302039 & 0.473986 & 0.382824     &          \\ \hline    
\end{tabular}
}
\caption{The agreement measures from tasks T1 and T2, according to language groups. Values above 0.6 are deemed substantial, and above 0.8 almost perfect agreement (shown in bold). We see that F-Kappa plummets down on T1.3 and T2.1, while R-Kappa keeps closer to K-Alpha, signalling possible cases of high agreement and low kappa paradoxes due to class imbalance. As shown in Section~\ref{sec:aggreg}, these are very imbalanced tasks, having classes with less than $1\%$ frequency.}
\label{tab:agreement}
\vspace*{-12pt}
\end{table}

\begin{table}[h]
\centering
{\small
\renewcommand{\arraystretch}{0.9}
\vspace*{0pt}
\begin{tabular}{|l|l|l|l|l|}
\hline
Question                         & URL Type          & F-Kappa     & R-Kappa  & K-Alpha \\ \hline
\multirow{2}{*}{T1.1}            & \textbf{External} & \textbf{0.545904} & 0.709033          & \textbf{0.578542}    \\ \cline{2-5} 
                                 & Internal          & 0.419651          & \textbf{0.781818} & 0.442124             \\ \hline
\multirow{2}{*}{T1.2}            & \textbf{External} & NA                & NA                & \textbf{0.638910} \\ \cline{2-5} 
                                 & Internal          & NA                & NA                & 0.561894             \\ \hline
\multirow{2}{*}{T1.3}            & \textbf{External} & \textbf{0.496923} & \textbf{0.658333} & \textbf{0.473396}    \\ \cline{2-5} 
                                 & Internal          & -0.059503         & 0.266667          & 0.219543             \\ \hline
\multirow{2}{*}{T2.1}            & External          & 0.481512          & 0.573828          & 0.776471             \\ \cline{2-5} 
                                 & \textbf{Internal} & \textbf{0.533927} & \textbf{0.746032} & \textbf{0.796330}    \\ \hline
\multirow{2}{*}{T2.2}            & \textbf{External} & \textbf{0.621433} & \textbf{0.664972} & \textbf{0.717461}    \\ \cline{2-5} 
                                 & Internal          & 0.538978          & 0.610342          & 0.631743             \\ \hline
\multirow{2}{*}{T2.3}            & \textbf{External} & \textbf{0.532366} & \textbf{0.582883} & \textbf{0.622741} \\ \cline{2-5} 
                                 & Internal          & 0.456065          & 0.518554          & 0.550125             \\ \hline
\end{tabular}
}
\caption{The agreement measures from tasks T1 and T2, according to URL types. In bold, the higher scores when comparing external vs internal URL types.}
\label{tab:agreement_url}
\vspace*{-12pt}
\end{table}

\subsubsection{Aggregated Annotations}
\label{sec:aggreg}

Each reference node was annotated by five workers on both tasks \textbf{T1} and \textbf{T2}. For all questions except \textbf{T1.2}, we used majority voting to aggregate the five annotations, and used the median for \textbf{T1.2}. For questions \textbf{T1.3}, and all of \textbf{T2}, ties could happen and were decided randomly. The percentage of ties out of all annotations was $15\%$ for \textbf{T1.3}, which is understandable given that different workers might have different access issues. Across all three \textbf{T2} sub-tasks, we had only a small share of ties, in each case less than $3\%$.

After combining this with the automatically checked references and the golden standard references (as described in Section~\ref{subsubsec:methcrowdaggr}), we reached the final dataset, whose annotation distribution can be seen in Table~\ref{tab:crowdaggr}. The table shows responses to questions T1 through T2.2. T2.3 describes the subtypes of publishers, which are 11 in total, so for a more informative and concise reporting we show the final authoritativeness classifications resulting from the combination of publisher subtypes (T2.3) and author types (T2.1). The table used for this classification can be found in Appendix~\ref{appendix:A}.

As seen in Table~\ref{tab:crowdaggr}, all languages have around $90\%$ of relevant references, a high number. The majority of references are also fairly to very easy to access. English is considerably harder to navigate than the other language groups, mainly due to lots of references pointing towards hard-to-interpret JSON objects, scientific, and biology databases. Spanish, Portuguese and Dutch all have a staircase pattern, with very easy references being the most common and harder references progressively rarer. Japanese would follow that too if not for the large number of references to VIAF, which are mostly classified as a $2$ on the Likert scale. Swedish is a more complex case, due to its main bulk of references being to Wikimedia's monuments tool, which is mostly a $3$ on the scale, heavily skewing the distribution. English is also the language with the most variation of barriers, needing domain knowledge $5.88\%$ of the time, and presenting security risks $11.76\%$ of the time. The domain knowledge issue was expected, and the security risks are likely due to a specific website domain used in many references, which does not use SSL certificates. 

Most language groups seem to have similar distributions of author types, with individuals being very rare, organisations being the majority, and collectives varying in-between. Swedish, however, has its references coming mainly from Wikimedia projects. Spanish, Portuguese, and Japanese also make vast use of Wikipedia articles as references. Publisher types vary between language groups. English has over $83\%$ of its references coming from academic and scientific publishers, which was expected. Spanish has Wikipedia as its main source of references, while Portuguese has the best balance between all types. Swedish and Japanese are both almost solely either Wikimedia projects or organisations. Dutch is populated mainly by museums, libraries, art, and heritage centres. This is reflected in the final scores for authoritativeness of each language too. English and Dutch are mainly authoritative, with other languages falling behind due to the number of references to Wikimedia projects.

According to feedback from crowd workers, when confronted with JSON websites they could not interpret, or when redirects happened, they would mark the 'Could not access' option. This means that this option gathers all information access problems under the same umbrella, which is a limitation of our study that could be further explored.

\begin{table}[]
\centering
{\small
\setlength{\tabcolsep}{1.8pt}
\renewcommand{\arraystretch}{0.9}
\begin{tabular}{|l|r|r|r|r|r|r|r|r|r|r|r|r|r|}
\cline{2-14}
\multicolumn{1}{l}{~}  & \multicolumn{1}{|l}{T1.1} &   \multicolumn{5}{|l|}{T1.2} &   \multicolumn{7}{l|}{T1.3} \\ \hline
Language &   R       &   0      &   1      &   2       &   3       &   4       &   0       &   1      &   2      &   3      &   4      &   5       &   6 \\ \hline
All  &   \parbox{24pt}{\raggedleft 91.73\%} &   3.78\% &   2.88\% &   14.91\% &   34.26\% &   44.17\% &   19.37\% &   2.09\% &   0.52\% &   0.52\% &   1.05\% &   51.31\% &   25.13\% \\ \hline
Dutch &   87.27\% &   0.30\% &   0.60\% &   6.85\% &   14.88\% &   77.38\% &   6.12\%  &   0.00\% &   0.00\% &   0.00\% &   0.00\% &   61.22\% &   32.65\% \\ \hline
English &   91.17\% &   19.37\%&   8.26\% &   2.28\% &   59.26\% &   10.83\% &   5.88\%  &   11.76\%&   0.00\% &   2.94\% &   5.88\% &   47.06\% &   26.47\% \\ \hline
Japanese &   92.73\% &   0.00\% &   0.84\% &   44.82\% &   11.20\% &   43.14\% &   64.29\% &   0.00\% &   0.00\% &   0.00\% &   0.00\% &   25.00\% &   10.71\% \\ \hline
Portuguese &   91.17\% &   1.14\% &   5.70\% &   17.38\% &   30.20\% &   45.58\% &   11.76\% &   0.00\% &   2.94\% &   0.00\% &   0.00\% &   73.53\% &   11.76\% \\ \hline
Spanish &   93.25\% &   0.28\% &   0.84\% &   12.26\% &   36.21\% &   50.42\% &   15.38\% &   0.00\% &   0.00\% &   0.00\% &   0.00\% &   50.00\% &   34.62\% \\ \hline
Swedish &   94.81\% &   1.64\% &   1.10\% &   5.48\%  &   52.60\% &   39.18\% &   30.00\% &   0.00\% &   0.00\% &   0.00\% &   0.00\% &   35.00\% &   35.00\% \\ \hline
\end{tabular}

\setlength{\tabcolsep}{1.8pt}
\hspace*{1.5pt}\begin{tabular}{|l|r|r|r|r|r|r|r|r|r|r|r|r|r|}
\cline{2-14}
\multicolumn{1}{l|}{} &   \multicolumn{4}{l|}{T2.1} &   \multicolumn{6}{l|}{T2.2.} &   \multicolumn{3}{l|}{Auth.} \\ \hline
\multicolumn{1}{|l|}{Language} & I. & O. & C. & N. & A. & C. & G. & Ne. & S. & N. & Yes & No & Ina. \\ \hline
\multicolumn{1}{|l|}{All}  & \parbox{24pt}{\raggedleft 0.69\%} & 67.66\% & 28.87\% & 2.77\% & 22.16\% & 37.75\% & 6.71\% & 1.77\% & 28.79\% & 2.81\% & 66.97\% & 30.22\% & 2.81\% \\ \hline
\multicolumn{1}{|l|}{Dutch} & 0.26\% & 94.81\% & 4.68\% & 0.26\% & 8.75\% & 71.43\% & 14.81\% & 0.52\% & 4.42\% & 0.26\% & 94.81\% & 4.94\% & 0.26\% \\ \hline
\multicolumn{1}{|l|}{English} & 0.52\% & 91.43\% & 5.19\% & 2.86\% & 83.12\% & 5.19\% & 3.64\% & 0.26\% & 4.94\% & 2.86\% & 90.91\% & 6.23\% & 2.86\% \\ \hline
\multicolumn{1}{|l|}{Japanese} & 0.00\% & 53.51\% & 42.08\% & 4.42\% & 1.30\% & 48.57\% & 2.60\% & 0.78\% & 42.34\% & 4.42\% & 52.47\% & 43.12\% & 4.42\% \\ \hline
\multicolumn{1}{|l|}{Portuguese} & 2.60\% & 67.53\% & 29.87\% & 0.00\% & 17.92\% & 31.95\% & 13.51\% & 6.75\% & 29.61\% & 0.26\% & 65.71\% & 34.03\% & 0.26\% \\ \hline
\multicolumn{1}{|l|}{Spanish} & 0.78\% & 55.84\% & 34.29\% & 9.09\% & 21.56\% & 27.79\% & 5.19\% & 2.08\% & 34.29\% & 9.09\% & 55.32\% & 35.58\% & 9.09\% \\ \hline
\multicolumn{1}{|l|}{Swedish} & 0.00\% & 42.86\% & 57.14\% & 0.00\% & 0.52\% & 41.56\% & 0.52\% & 0.26\% & 57.14\% & 0.00\% & 57.40\% & 42.60\% & 0.00\% \\ \hline
\end{tabular}
}
\caption{Distribution of reference nodes according to aggregated crowdsourced responses. On T1.1, R = Relevant. On T1.2, each number is a level on the likert scale. On T1.3, each number represents a barrier as described in Appendix~\ref{sec:appA_T1}. On T2.1, I = Individual, O = Organisation, C = Collective, N = No access. On T2.2, A = Academic/Scientific, C = Company/Organisation, G = Government, Ne = News, S = Self-published, N = No access. Auth gives the final authoritativeness scores for Yes, No, and Inaccessible.}
\label{tab:crowdaggr}
\vspace*{-12pt}
\end{table}

\subsection{Machine Learning}
\label{subsec:evalml}


To reduce the influence of overfitting in our evaluation, we split the data into training and testing by following a 10-fold stratified cross-validation. Scores from each fold were macro averaged, in consideration of the heavy class imbalance (see Section~\ref{subsubsec:methML}). To compare results between model configurations, and across distinct algorithms, we used the F1-score, also due to the class imbalance that is present on all three tasks.

\begin{table}[h]
\centering
{\small
\setlength{\tabcolsep}{1.8pt}
\renewcommand{\arraystretch}{0.9}
\begin{tabular}{|l|r|r|r|r|l|l|r|r|r|r|l|l|r|r|r|r|}
\cline{1-5} \cline{7-11} \cline{13-17}
\multicolumn{5}{|l|}{Relevance}                                                                                                                      &  & \multicolumn{5}{l|}{Authoritativeness}                                                                               &  & \multicolumn{5}{l|}{Ease of Navigation}                                                                              \\ \cline{1-5} \cline{7-11} \cline{13-17} 
Model & \multicolumn{1}{l|}{Acc.}           & \multicolumn{1}{l|}{Prec.}          & \multicolumn{1}{l|}{Rec.}  & \multicolumn{1}{l|}{F1}             &  & Model & \multicolumn{1}{l|}{Acc.} & \multicolumn{1}{l|}{Prec.} & \multicolumn{1}{l|}{Rec.} & \multicolumn{1}{l|}{F1} &  & Model & \multicolumn{1}{l|}{Acc.} & \multicolumn{1}{l|}{Prec.} & \multicolumn{1}{l|}{Rec.} & \multicolumn{1}{l|}{F1} \\ \cline{1-5} \cline{7-11} \cline{13-17} 
RFC & 0.936 & 0.796 & 0.772 & 0.780 &  & RFC & 0.983 & 0.983 & 0.979 & 0.981 &  & RFC & 0.834 & 0.825 & 0.824 & 0.815 \\ \cline{1-5} \cline{7-11} \cline{13-17} 
XGB   & 0.918                               & 0.738                               & \textbf{0.8}               & 0.762                               &  & XGB   & 0.980                     & 0.978                      & 0.976                     & 0.977                   &  & XGB   & 0.060                     & 0.185                      & 0.367                     & 0.183                   \\ \cline{1-5} \cline{7-11} \cline{13-17} 
GNB   & 0.865                               & 0.641                               & 0.731                      & 0.668                               &  & GNB   & 0.977                     & 0.979                      & 0.969                     & 0.974                   &  & GNB   & 0.806                     & \textbf{0.871}             & 0.781                     & 0.811                   \\ \cline{1-5} \cline{7-11} \cline{13-17} 
SVM   & 0.859                               & 0.651                               & 0.783                      & 0.683                               &  & SVM   & 0.979                     & 0.980                      & 0.971                     & 0.975                   &  & SVM   & 0.824                     & 0.760                      & 0.829                     & 0.780                   \\ \cline{1-5} \cline{7-11} \cline{13-17} 
NN    & \multicolumn{1}{l|}{\textbf{0.943}} & \multicolumn{1}{l|}{\textbf{0.835}} & \multicolumn{1}{l|}{0.766} & \multicolumn{1}{l|}{\textbf{0.791}} &  & NN    & \textbf{0.984}                     & \textbf{0.985}                      & \textbf{0.979}                     & \textbf{0.982}                   &  & NN    & \textbf{0.936}            & 0.848                      & \textbf{0.830}            & \textbf{0.831}          \\ \cline{1-5} \cline{7-11} \cline{13-17} 
\end{tabular}
}
\caption{Results from the 3 ML classification tasks: relevance, authoritativeness, and navigation effort. Reported metrics are accuracy, macro averaged precision, recall and F1-score.}
\label{tab:ML}
\vspace*{-12pt}
\end{table}

The best results from each model, in each of the three tasks, are shown in Table~\ref{tab:ML}. We compare results from the best feature and hyperparameter configurations employed in each of the five model algorithms, namely random forest classifiers (RFC), gradient boosted decision trees (XGB), gaussian naive Bayes (GNB), support vector machines with a radial basis function kernel (SVM), and neural networks (NN).

Most algorithms show modest to very good results in each of the three tasks, confirming that our reasoning to choose them was justified, with NNs outperforming the others in all tasks. Differences in feature engineering and hyperparameter tuning resulted in F1-score results that varied as little as 1\% for NNs on the authoritativeness task and as much as 63\% for SVMs in the ease of navigation task. The full list of model configurations tested and metrics obtained can be found in the GitHub repository mentioned in Section~\ref{sec:introduction}. Here we will describe the best performing models and discuss some of the configuration choices that had the most perceived impact.

\subsubsection{Relevance task}
On modelling relevance, i.e. given a reference and its associated claim, predict either the class \textit{irrelevant} or the class \textit{relevant}. The best configuration was as follows: we scaled each feature so that their maximum absolute values would be 1.0, one-hot encoded all categorical features, using a \textit{other} category with a 1\% frequency threshold, did not employ PCA, and dropped all features whose correlation coefficient was higher than 95\%. The data was then resampled with BorderlineSMOTE. The network consisted of two fully connected layers with a leaky ReLu activation and a dropout layer with 50\% probability, trained to minimise categorical cross-entropy via an Adam optimiser with learning rate set to 1e-3.

\subsubsection{Authoritativeness task}
On modelling authoritativeness, i.e. given a reference and its source, predict either the class \textit{authoritative} or the class \textit{not authoritative}, NNs performed reasonably better than other options. It is worth noting, however, that they have much more hyperparameters and took much more time to train than random forest classifiers, which scored extremely closely. Here, we ended up with exactly the same network setup, loss, and optimiser as for the relevance task. The feature engineering was different: knowledge graph embeddings without PCA and without dropping any features from correlation yielded the best performance.

\subsubsection{Ease of navigation task}
On modelling ease of navigation, i.e. given a reference and its source, predict one out of five classes which describe the level of difficulty of navigating the source, NNs again performed the best. The network setup, loss, and optimiser were again the same, and one-hot encoding without PCA and without dropping features was used. SMOTE oversampling was used.

\subsubsection{Model and encoding comparisons}
While NNs scored higher, one should be cautious of the possibility of overfitting. Random forests are smaller, faster to train, more robust to overfitting, and scored relatively close to NNs overall, making them a fair option. XGB, being a boosting technique, probably did not handle the dimensionality as well as RFC. Naive-Bayes assumes features are independent, which is hardly the case even after dropping correlated features, helping justify its lower performance. SVMs work well when data can be clearly separated, even in high dimensional spaces, leading us to conclude that the data must have a noticeable degree of overlapping the kernel did not manage to address.

Encoding options seemed to have an impact on average performance across model configurations depending on the used algorithms. Both RFC and XGB with label encoding performed similar to or better than one-hot encoding or embeddings on all tasks, which was to be expected given both are based on decision trees. For all other models, on all tasks, embeddings and one-hot encoding both performed much better than label encoding, with one-hot encoding performing the best, especially in NNs.

PCA had a slightly negative effect on both RFC and XGB, but positive overall on other algorithms, which is to be expected as decision trees handle high dimensionality very well in comparison to other algorithms. SMOTE was the best oversampling technique on all algorithms and tasks by a large margin. Lastly, dropping correlated features had mixed results depending on the model and task.
\section{Discussion}
\label{sec:discussion}

\subsection{Improving provenance representation}

We detected considerable variation in quality depending on how the reference node encodes its source: external URLs, internal Wikidata entities, Wikipedia links, etc. Wikipedia import URLs are highly relevant ($98\%$), more than external URLs ($90\%$) and considerably more than stated references ($26\%$), the two most frequent types. Wikipedia import URLs are also very easy to access, with $77\%$ of relevant references falling on level $4$ of the Likert scale used in question T1.2, much more than external URLs ($38\%$ on level 4) and stated URLs ($0\%$ on level 4, with $50\$$ on level 0). Finally, non-relevant Wikipedia references were only deemed so due to errors in their contents, while non-relevant external references, for instance, were often unavailable ($24\%$).

Wikimedia projects, e.g. Wikipedia, generally provide relevant and easy to access references. However, these sources are inherently not authoritative. The reason for that is not only their collective nature but also that they are secondary sources. According to the Wikidata guidelines~\cite{wikidataguidelines}, references must be primary sources. The Wikidata community has made efforts through the years to replace and avoid this type of references, opting instead for sources that may be authoritative but also be less relevant and harder to navigate. To facilitate the community's work, possible approaches may replace e.g. references pointing to Wikipedia pages by automatically extracting relevant and authoritative citations within them. 

Our data also shows how the use of \texttt{stated in} predicates without the use of direct external URLs or identifiers creates references of poor quality; they are very likely irrelevant, hard to access, or unavailable. Wikidata editors thus should provide a direct URL or external identifier whenever possible, and this should be mentioned in official guidelines.

\subsection{Wikidata's content across languages}

Quality measures from references in Wikidata differ across languages, as well as agreement from the crowd. We believe this is mainly tied to the very distinct types of content that are normally linked to references in each language. For instance, English references have a high percentage ($71\%$) of links to JSON objects and mainly pertain to biology and/or scientific data ($83\%$). They present much lower ease of access ($18\%$ on level 0) than other languages and are the only ones that might impose barriers to access in the form of required domain knowledge. Similarly, Dutch references link mostly to cultural entities such as museums ($65\%$), Swedish's most referenced domain ($42\%$) is the monument heritage tool \footnote{\url{https://heritage.toolforge.org/}} from Wiki Loves Monuments, and Japanese references rely heavily on Wikipedia ($41\%$).

This shows how much power individuals and organised groups can have over Wikidata's overall quality. One example is WikiProject Astronomy, which oversees the insertion and maintenance of entities relating to astronomy, which are nowadays over 7M in number ($7.6\%$ of all Wikidata entities). Wikidata's contents are thematically skewed in specific languages, and might also be so in its entirety. Whether or not Wikidata will remedy this in the future, further analysis of its data must take this imbalance into account, i.e. by exploring thematic clusters in isolation or by excluding the largest ones from the analysis.

Based on our findings we recommend the following improvements for Wikidata:
\begin{itemize}[label={}]
  \item \textbf{Reference quality}: Perform automated checks to point users to problematic references, highlighting the type of issue -- i.e. relevance, authoritativeness, or ease of access -- that is likely to affect them.
  \item \textbf{Missing external URLs}: Explicitly mention in the guidelines to provide a direct URL or external identifier whenever possible; implement automated prompts to do so in the user interface.
  \item \textbf{Wikimedia sources}: Actively support users in replacing Wikimedia sources, e.g. by implementing an approach to automatically. extract citations from Wikipedia pages.
  \item \textbf{Differences across languages}: Encourage contributions from users from different backgrounds, e.g. with initiatives in the fashion of WikiProjects, in order to make quality and coverage of references more homogeneous across languages.
\end{itemize}

%

\subsection{ML results, features, and the use of embeddings} 

The main goal of the ML models built in this paper was to provide a better understanding of the automation of the tasks described in Section~\ref{subsubsec:methML}, and to define baselines for them. These are essential for the assessment of the relevance of the text referred to by the claim. 

To our knowledge, this hybrid crowdsourced/automated model for checking knowledge graph references has never been done before in the literature. 
The results obtained by our models indicate great potential for assisted editing, for other ML tasks that rely on surface features of the underlying data structure, and for models that use content-related features. This applies not only to Wikidata but potentially to other collaborative knowledge graphs as well.

Finally, knowledge graph embeddings were the most promising of the different model configurations applied. However, they did not perform better than the much simpler one-hot encoding. We believe this is in part due to dimensionality and sparsity being greatly raised and the lack of sufficient training data to model it. It is also because embeddings leverage similarities between ontological entities, yet our data does not have a large enough variety of entities to properly benefit from them. This shows that for some scenarios and tasks the usability of knowledge graph embeddings is limited and other encoding methods should be preferred.

\subsection{Limitations}

There are limitations to be taken into account in this study, namely two potential sources of bias: the crowd workers and the data volatility of Wikidata.

Crowd workers were taken from MTurk, a platform where demographics of users, such as gender, nationality, education and familiarity with web technologies is likely not the same as that of Wikidata editors, which is the target group of this study. This bias should be considered when analysing the perceived quality measures reported in this paper, in comparison to what is perceived by the Wikidata community.

The version of Wikidata analysed here is from a given point in time, and the changing nature of Wikidata must be considered. Our work shows the current strengths and weaknesses of Wikidata, provides valuable information to remedy faults, and, most importantly, defines a process that can be repeatedly carried to observe how Wikidata evolves, allowing the community to respond accordingly.

\section{Conclusion}
\label{sec:conclusion}

In this paper, we explored several questions surrounding the quality of Wikidata references, following the Wikidata guidance on the expected standard for references. We focused on three key dimensions: relevance, ease of access, and authoritativeness. We have shown statistics of how Wikidata organises its references, and have measured their quality through a hybrid computational workflow, with the use of automated API calls and crowdsourcing. Furthermore, we have investigated how different language groups and reference node types impact the measured quality. Finally, we have created a training dataset that was used to model the classification problems which are important for helping Wikidata automate its reference curation process. We showed this is feasible by training both traditional and deep learning ML models on the dataset with promising results.

This is the first study to tackle how the quality of Wikidata varies according to language and reference type. It is important to know whether Wikidata's efforts of improving the quality of its contents have been having effects on the experience of the whole of its user community or only in its majority. We also uncovered which forms of reference encoding work well and which do not. Both these elements directly expand on the work previously done by us in \cite{Piscopo2017}.

Although modelling relevance and authoritativeness of Wikidata references has already been done, it previously has made use of features coming from the edit history. On our work, we use solely the state of references and its URLs as they are. It showed encouraging results and will help guide us and others on future work. This will look into the reasons why some of the discrepancies of measured quality occur between languages or reference types, and why workers have a harder time agreeing on measurements for specific types of references. Further studies could also look at the generated dataset, which can be valuable for misinformation detection tasks. As steps to take next, we aim to process the feedback and comments left by workers of this crowdsourcing campaign to understand the reasons why they have deemed references as either irrelevant or inaccessible. We will also look into different forms of automating the assessment of reference relevance, by extracting semantic information directly from the reference's content.

\paragraph{Acknowledgements} The project leading to this publication has received funding from the European Union’s Horizon 2020 research and innovation programme under the Marie Sklodowska-Curie grant agreement No. 812997 (Cleopatra).
\vspace*{-6pt}
\bibliographystyle{unsrt} 
\bibliography{bibliography}

\newpage

\appendix

\section{Description of Crowdsourcing Tasks}
\label{appendix:A}

\subsection{Recruitment.}


Crowdsourcing was carried through Amazon's Mechanical Turk platform.\footnote{\url{https://www.mturk.com/}} Mechanical Turk is a marketplace-like platform, where users may register as anonymous 'workers', who can browse a list of tasks, select one to work on, and get paid for their work.

Our tasks have a standard structure and approach to welcoming workers. First, they are shown a sequence of introductory pages:

\begin{enumerate}
    \item Page $1$ provides them with a brief overview of the project, including what their contributions will be used for, stating any language requirements the task may ask for, and information on ethics approval and university contract;
    \item Page $2$ explains what the task will be and what actions they will need to perform;
    \item Page $3$ shows them useful examples that will help with finishing the task better and faster;
    \item Page $4$ lists a set of rules which the users have to keep in mind during the task.
\end{enumerate}

The examples given at introductory page ($3$) are mock values that resemble those that could be shown to the user during the task, along with illustrative images and text describing what a correct response to that task would be and why. These rules shown at introductory page ($4$) hint at the quality assurance techniques, which are specified further below, e.g. stating that workers should fill all information requested, that their contributions will be checked for spam and that they have a limit of retries for the task. The introductory pages can be quickly skipped, in the case of returning workers, and accessed again at any point during the task for clearing doubts.

After the introductory pages, workers are shown a small randomised language quiz, which serves quality assurance purposes and is further described in Section~\ref{paragraph:QA}.

\subsection{Design of Task T1}
\label{sec:appA_T1}

After browsing through the instructions, including the examples, and passing the language test, the worker is shown the first of a set of six micro-tasks contained in that task.

\begin{figure}[]
  \centering
  \includegraphics[width=0.7\linewidth]{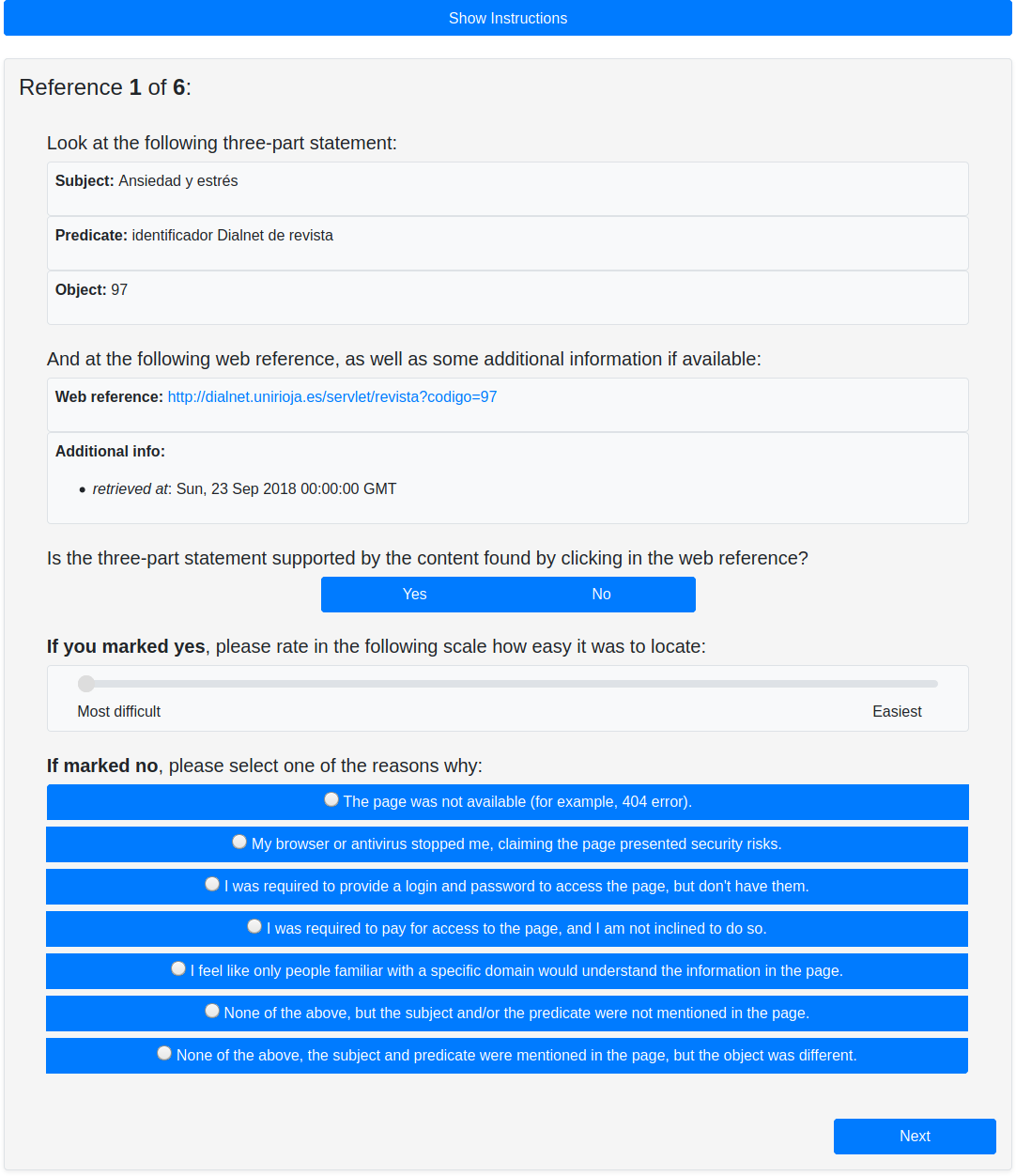}
  \caption{The design of the first task (\textbf{T1}), as seen by workers.}
  \label{fig:task1}
\end{figure}

The exact design of T1 is shown in Fig.~\ref{fig:task1}. Workers are presented with a small set of reference nodes to be judged, taken from the unautomatable portion of the $2310$ references sampled across the six target languages. Each reference node is displayed in a user-friendly manner, as a combination of its extracted representative URL, its predicates and their objects, and a claim node that is linked to it.

Both nodes are shown to the worker in a friendly manner. Starting from the top of the page, the worker sees a numbering of the micro-tasks they are currently working on. Below it, the statement node is shown as a subject, predicate, and object triple, the interpretation of which has been explained in the instructions. Following, they are shown the reference node, in the form of a URL pointing to a web resource, and additional info found in that node. The URL is obtained through multiple methods depending on the form of the reference node and is described in Section~\ref{subsec:data}.

The questions T1.1, T1.2 and T1.3 are described in Section~\ref{subsubsec:methcrowd}. On T1.2, each step of the scale has a description which states what would be a suitable example of a reference falling there. They are, from least to most accessible:
\begin{itemize}
    \item `I had to navigate the website and use the additional information or infer the statement using my common sense';
    \item `I had to navigate the website and use the additional information to find it';
    \item `I had to navigate the website to find it';
    \item `I did not need to navigate the website but had to read through the content to find it';
    \item `I did not need to navigate the website or read through much content to find it'.
\end{itemize}

Where, in the instructions, we describe navigating the website as interacting with the website controls or moving through pages while in the same domain, while looking for the information.

On T1.3, the barriers we listed as options are:
\begin{enumerate}
    \item \textbf{The page was not available}: If a $404$ error code, redirection to home page or wrong page is detected.
    \item \textbf{Security issues}: If the worker's browser stops them from accessing the page, claiming security risks.
    \item \textbf{Credentials needed}: In case login and password or special permission is needed to access the website or part of it.
    \item \textbf{Paywall}: In case the worker is asked to pay for access to the website or part of it.
    \item \textbf{Domain knowledge}: If the worker feels like the website requires domain knowledge in order to understand its content or part of it.
    \item \textbf{Subject or predicate not mentioned}: In case either subject or predicate is missing from the page, which indicates that the claim information is not found there and thus the reference is just not relevant.
    \item \textbf{Subject and predicate mentioned, but not object}: In this case, both subject and predicate are found, but the object is different, which indicates that not only the reference is not relevant, but might contradict the claim.
\end{enumerate}

\subsection{Design of Task T2}

To define authoritativeness, we first define a reference's author and publisher types. We use three author types: Individual, where the author is an identifiable person; Organization; and Collective, where the author is the combined effort of multiple anonymous users. We use five publisher types, some with sub-types. Table~\ref{tab:task2types} shows the types of authors and publishers, as well as whether or not their combination is likely to express authoritativeness. We use this table when assessing the authoritativeness of web-pages in T2.

\begin{figure}[]
  \centering
  \includegraphics[width=0.7\linewidth]{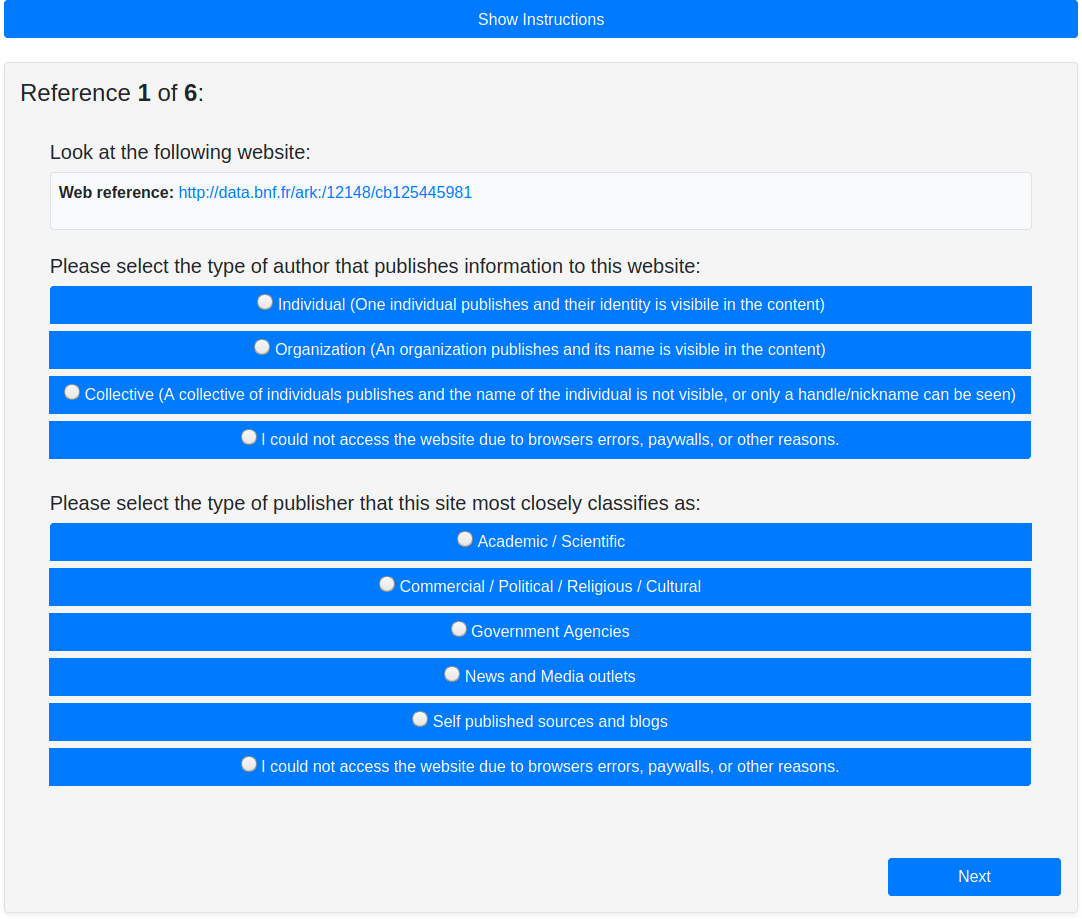}
  \caption{The design of the second task (\textbf{T2}), as seen by workers.}
  \label{fig:task2}
\end{figure}

\begin{table}[]
\centering
\small
\renewcommand{\arraystretch}{0.9}
\begin{tabular}{|l|l|ccc|}
\hline
\textbf{Publisher type} &
  \textbf{Publisher sub-type} &
  \multicolumn{3}{c|}{\textbf{Author}}
   \\ \cline{3-5} 
 &
   &
  \multicolumn{1}{c|}{I} &
  \multicolumn{1}{c|}{O} &
  C \\ \hline
\begin{tabular}[c]{@{}l@{}}Academic and scientific\\  organizations\end{tabular} &
  Academic and research institutions &
  \multicolumn{1}{c|}{Yes} &
  \multicolumn{1}{l|}{Yes} &
  No \\ \cline{2-5} 
 &
  Academic publishers &
  \multicolumn{1}{c|}{Yes} &
  \multicolumn{1}{l|}{Yes} &
  No \\ \cline{2-5} 
 &
  Other academic organizations &
  \multicolumn{1}{c|}{Yes} &
  \multicolumn{1}{l|}{Yes} &
  No \\ \hline
\begin{tabular}[c]{@{}l@{}}Companies and non-academic/\\ scientific organizations\end{tabular} &
  Vendors and e-commerce companies &
  \multicolumn{1}{c|}{Yes} &
  \multicolumn{1}{l|}{Yes} &
  No \\ \cline{2-5} 
 &
  Political or religious organisations &
  \multicolumn{1}{c|}{No} &
  \multicolumn{1}{l|}{No} &
  No \\ \cline{2-5} 
 &
  Cultural institutions &
  \multicolumn{1}{c|}{No} &
  \multicolumn{1}{l|}{No} &
  No \\ \cline{2-5} 
 &
  Other types of company &
  \multicolumn{1}{c|}{Yes} &
  \multicolumn{1}{l|}{Yes} &
  No \\ \hline
\begin{tabular}[c]{@{}l@{}}Government agencies\\ and authorities\end{tabular} &
  - &
  \multicolumn{1}{c|}{Yes} &
  \multicolumn{1}{l|}{Yes} &
  No \\ \hline
News and media outlets &
  \begin{tabular}[c]{@{}l@{}}Traditional news and media\\ (e.g. news agencies, broad- casters)\end{tabular} &
  \multicolumn{1}{c|}{Yes} &
  \multicolumn{1}{l|}{Yes} &
  No \\ \cline{2-5} 
 &
  \begin{tabular}[c]{@{}l@{}}Non-traditional news and media\\ (e.g. online magazines, platforms to\\ collaboratively create news)\end{tabular} &
  \multicolumn{1}{c|}{Yes} &
  \multicolumn{1}{l|}{No} &
  No \\ \hline
Self-published sources &
  - &
  \multicolumn{1}{c|}{No} &
  \multicolumn{1}{l|}{No} &
  No \\ \hline
\end{tabular}
\caption{The types and sub-types of publishers, along with the types of authors: \textit{I} stands for Individual, \textit{O} for Organization and \textit{C} for Collective.}
\label{tab:task2types}
\end{table}

T2 follows a similar structure to T1, as seen in Fig.~\ref{fig:task2}. Like in T1, after browsing through the instructions and language test, workers are shown a small set of reference nodes to be judged. The set is taken from the same unautomatable portion of the $2310$ references used in T1, with the exception of those pointing towards Wikipedia, as we discussed. Each reference node is displayed as its extracted representative URL only.

\paragraph{Task Time and Monetary Rewards}

Before running the actual crowdsourcing campaigns, we launched pilot versions of T1 and T2, to collect feedback on the task designs, and also to measure the time workers spent to finish the tasks. We then took the average time and used it to define appropriate monetary compensation for the tasks, by using the current US minimum wage of USD $7.25$. Considering that each task is to be assigned to five different workers to gather consensus, and that two of the six reference nodes in each task are in fact gold-standards, we arrive at the final cost for each non-gold-standard micro-task for both T1 and T2. This is seen in Table~\ref{tab:prices}.

In Section~\ref{subsec:evalcrowdsourcing} of the paper, we show the time crowd workers spent on the actual tasks. The mean time spent by workers on a single annotation was $36$ seconds. Using the reasoning just described, this equals to a due average payment of USD $0.43$ per full task, which agrees with what we calculated for the pilot. Every worker was paid at least USD $0.5$ per task, with workers for the Swedish and Japanese groups being paid USD $0.8$ per task; this was due to their rarity in the platform, meaning we had to increase the reward to speed up task completion.

\begin{table}[]
\centering
\small
\renewcommand{\arraystretch}{0.9}
\begin{tabular}{|l|l|l|l|l|}
\hline
 &
  \begin{tabular}[c]{@{}l@{}}Avg. time per task\\ (seconds)\end{tabular} &
  \begin{tabular}[c]{@{}l@{}}Payment per\\ task (USD)\end{tabular} &
  \begin{tabular}[c]{@{}l@{}}Number of worker\\ assignments\end{tabular} &
  \begin{tabular}[c]{@{}l@{}}Total cost per non \\gold-standard micro-task\end{tabular} \\ \hline
T1 &
  230.14 &
  0.47 &
  5 &
  0.58 \\ \hline
T2 &
  223.91 &
  0.45 &
  5 &
  0.56 \\ \hline
\end{tabular}
\caption{Table with the average times and costs of tasks for T1 and T2.}
\label{tab:prices}
\end{table}

\paragraph{Ethics}
As for any experiment carried with human participants, we have obtained ethics clearance for our crowdsourcing tasks. This was received on the $6$th of May of $2020$ by Kings College London, with registration number \textit{MRA-19/20-18878}. No personal and identifiable data was collected from the workers during the crowdsourcing tasks.
\newpage
\section{Features used on ML models}
\label{appendix:B}

\begin{table}[h]
\centering
\small
\setlength{\tabcolsep}{1.8pt}
\renewcommand{\arraystretch}{0.9}
\begin{tabular}{|p{3cm}|p{10cm}|}
\hline
\multicolumn{2}{|l|}{\textbf{Features of the representative URL extracted}}                                                                                                                                                                                                                                                                                \\ \hline
Type of extraction               & Whether the URL was a direct \textit{reference URL}, a \textit{stated in} reference, an external identifier, an URL imported from a Wikimedia project, or inferred from another item.                                                                                                 \\ \hline
URL suffix                       & The URL's suffix. This is extracted as being whatever lies between the top-level-domain (tld) and the domain. For example, \textit{www.ebi.ac.uk}, \textit{ac} would be the suffix, while \textit{uk} would be the tld, and \textit{ebi} the domain. \\ \hline
URL scheme                       & Whether the URL uses HTTP or HTTPS.                                                                                                                                                                                                                                                                                     \\ \hline
Top-level-domain                 & The URL's top-level-domain, e.g. uk, gov, org, etc.                                                                                                                                                                                                                                                                     \\ \hline
URL subdomain                    & The URL's subdomain, e.g. tools, api, www, etc.                                                                                                                                                                                                                                                                         \\ \hline
URL domain                       & The URL's domain, e.g. bbc, yahoo, ebi, etc.                                                                                                                                                                                                                                                                            \\ \hline
\multicolumn{2}{|l|}{\textbf{Features on the reference node's coding and structure}}                                                                                                                                                                                                                                                                       \\ \hline
Is inferred                      & Whether or not the reference node says the claim was inferred from other Wikidata object.                                                                                                                                                                                                                               \\ \hline
Stated in somewhere              & Whether or not the reference node says the claim what stated in some real source.                                                                                                                                                                                                                                       \\ \hline
Instance of stated-in source     & If the reference was stated in some source (using predicate \textit{P248}), we collect the Wikidata item representing the class of the source's Wikidata object, linked to it via the \textit{instance of (P31)} property. When using embeddings, this is a sparse vector representing the item.                                                                                           \\ \hline
Has external identifier          & Whether or not the reference node links to an external identifier.                                                                                                                                                                                                                                                      \\ \hline
Predicate of external identifier & The Wikidata item representing what type of external identifier is used, which is denoted by the associated predicate, e.g. VIAF ID, BNF ID, etc. When sing embeddings, this is a sparse vector representing he item.                                                                                                                                                                                                      \\ \hline
Number of identifier URLs        & Number of representative URLs found by following external identifiers.                                                                                                                                                                                                                                                  \\ \hline
Has direct URL                   & Whether or not the reference node has a direct and explicit URL.                                                                                                                                                                                                                                                        \\ \hline
Number of Wikimedia import URLs  & Some reference nodes import their information from other Wikimedia projects and refer to it via specific predicates (\textit{P143}), and this counts the amount of import links.                                                                                                                       \\ \hline
Retrieval date                   & The retrieval date of the reference if there is one.                                                                                                                                                                                                                                                                    \\ \hline
Publication date                 & The publication date of the reference if there is one.                                                                                                                                                                                                                                                                  \\ \hline
\multicolumn{2}{|l|}{\textbf{Features of the website available through the representative URL}}                                                                                                                                                                                                                                                            \\ \hline
Language                         & The language crawled and identified on the website.                                                                                                                                                                                                                                                                     \\ \hline
Content type                     & The type of content found on the website, such as HTML, JSON, XML, plain text, etc.                                                                                                                                                                                                                                     \\ \hline
\multicolumn{2}{|l|}{\textbf{Features of the claim node associated to this reference node}}                                                                                                                                                                                                                                                                \\ \hline
Statement rank                   & The rank of the claim, which can be normal, preferred or deprecated.                                                                                                                                                                                                                                                    \\ \hline
Statement datatype               & The type of object of the statement, i.e. time, wikidata object, numeric, string, etc.                                                                                                                                                                                                                                  \\ \hline
Instance of statement property   & We collect the Wikidata item representing the statement's main predicate and, since there are too many of them, make use of its class, which it connects to via the \textit{instance of (P31)} relation. When using embeddings, this is a sparse vector representing said item.
\\ \hline
\end{tabular}
\caption{Features used on the ML models.}
\label{tab:features}
\end{table}

\end{document}